\documentclass[fleqn,10pt]{wlscirep}
\usepackage[utf8]{inputenc}
\usepackage[T1]{fontenc}

\usepackage{amsmath} 
\usepackage{amssymb}  
\usepackage{threeparttable}
\usepackage[]{algorithm}
\usepackage[noend]{algpseudocode}

\usepackage{amsmath,amsfonts}
\usepackage{array}
\usepackage{textcomp}
\usepackage{stfloats}
\usepackage{url}
\usepackage{verbatim}

\usepackage{graphics} 
\usepackage{arydshln}
\usepackage{graphicx}
\usepackage{multirow}
\usepackage{amssymb}  
\usepackage{threeparttable}
\usepackage{booktabs}
\usepackage{hhline}
\usepackage{soul,color}
\soulregister\cite7
\soulregister\ref7
\soulregister\pageref7
\usepackage{color, colortbl}
	
\definecolor{LightCyan}{rgb}{0.88,1,1}
\usepackage{tikz}

\usetikzlibrary{arrows}
\usetikzlibrary{shapes}

\usepackage{nomencl}
\makenomenclature

\usepackage{color}
\definecolor{applegreen}{rgb}{0.55, 0.71, 0.0}

\usepackage[]{xcolor}
\definecolor{orange}{RGB}{255, 128, 0}
\definecolor{purple}{RGB}{102, 0, 204}
\definecolor{blue}{RGB}{0, 51, 102}
\definecolor{reds}{RGB}{102, 0,51}

\newcommand{\argmin}{\operatornamewithlimits{argmin}}

\title{Residual Dynamics Learning for  Trajectory Tracking for Multi-rotor Aerial Vehicles}

\author[1,2,*]{Geesara Kulathunga}
\author[3]{Hany Hamed}
\author[4]{Alexandr Klimchik}
\affil[1]{Centre for Robotics and Mechatronics Components, Innopolis University, 420500, Russia}
\affil[2]{Lincoln Institute for Agri-Food Technology, University of Lincoln,  LN1, United Kingdom}
\affil[3]{Advanced Institute of Science and Technology (KAIST), Daejeon, 28210, South Korea}
\affil[4]{School of Computer Science, University of Lincoln,  LN6, United Kingdom}

\affil[*]{gkulathunga@lincoln.ac.uk}

\begin{abstract}
This paper presents a technique to model the residual dynamics between a high-level planner and a low-level controller by considering reference trajectory tracking in a cluttered environment as an example scenario. We focus on minimising residual dynamics that arise due to only the kinematical modelling of high-level planning. The kinematical modelling is sufficient for such scenarios due to safety constraints and aggressive manoeuvres that are difficult to perform when the environment is cluttered and dynamic. We used a simplified motion model to represent the motion of the quadrotor when formulating the high-level planner. The Sparse Gaussian Process Regression-based technique is proposed to model the residual dynamics.  Recently proposed Data-Driven MPC is targeting aggressive manoeuvres without considering obstacle constraints. The proposed technique is compared with Data-Driven MPC to estimate the residual dynamics error without considering obstacle constraints. The comparison results yield that the proposed technique helps to reduce the nominal model error by a factor of 2 on average. Further, the proposed complete framework was compared with four other trajectory-tracking approaches in terms of tracking the reference trajectory without colliding with obstacles. The proposed approach outperformed the others with less flight time without losing computational efficiency.
\end{abstract}
\begin{document}

\flushbottom
\maketitle
\nomenclature[01]{$B_z$}{a mapping matrix that projects the full state vector $\mathbf{z}$ onto the subspace of states relevant to  the residual dynamics}
\nomenclature[02]{$\mathbf{f}_{norm}$}{nominal system dynamics, it can be either kinematic model or dynamic model}
\nomenclature[03]{$\mathbf{f}_{est}$ }{augmented quadrotor motion model that comprises nominal dynamics and estimated residual dynamics  by Sparse Gaussian Process (SGP)}
\nomenclature[04]{$\mathbf{f}_{d}$ }{discrete augmented system dynamics}
\nomenclature[05]{$\mathbf{g}_1(\mathbf{w}), \mathbf{g}_2(\mathbf{w})$}{ the constraints enforced by system dynamics and obstacle for  Nonlinear Model Predictive Control (NMPC)}
\nomenclature[06]{ $\mathbf{g}_m$ }{residual dynamics model that is learnt using Sparse Gaussian Process}
\nomenclature[07]{$g(\mathbf{z})$ }{residual dynamics model that is learnt using Gaussian Process}
\nomenclature[08]{$\mathbf{p}_k$ }{the robot's current position $\mathbf{p}^{\mu}_k, \mu \in \{x,y,z\}$}
\nomenclature[09]{$\mathbf{p}_k$ }{the robot's current velocity $\mathbf{v}^{\mu}_k, \mu \in \{x,y,z\}$}
\nomenclature[10]{$\mathbf{u}^{ref}$ }{desired reference robot control input at time step k }
\nomenclature[11]{$\mathbf{u}_k$ }{ the robot's current control inputs $\textbf{u}_k \in \mathbb{R}^{n_u}$, influencing its linear velocity $\mathbf{v}_k \in \mathbb{R}^3$ in meters per second and angular velocity $\omega$}
\nomenclature[12]{$\bar{\mathbf{u}}_k$ }{the estimated control input at time step k by NMPC}
\nomenclature[12]{$\hat{\mathbf{u}}_k$ }{actual control input at time step k after applying the current control ($\bar{\mathbf{u}}_k$) to the robot}
\nomenclature[13]{$\mathbf{w}$}{the estimated state trajectories and corresponding control inputs ($[\bar{\textbf{u}}_{k}, \hdots, \bar{\textbf{u}}_{k+N-1}, \bar{\textbf{x}}_{k}, \hdots, \bar{\textbf{x}}_{k+N}]$) estimated from the NMPC} 
\nomenclature[15]{$\bar{\mathbf{x}}_k$ }{the estimated robot state vector at time step k by using NMPC} 
\nomenclature[14]{$\mathbf{x}_k$}{the robot's current state $\textbf{x}_k \in \mathbb{R}^{n_x}$, given by its position $\mathbf{p}_k \in \mathbb{R}^3$ in meters and orientation $\alpha$ in radians at time step $k$}
\nomenclature[16]{$\hat{\mathbf{x}}_k$ }{actual robot state vector at time step k after applying the current control ($\bar{\mathbf{u}}_k$) to the robot}
\nomenclature[16]{$\mathbf{x}_m$}{ the optimal inducing points employed to construct an approximate multivariate Gaussian distribution by using SGP}
\nomenclature[17]{$\mathbf{x}^{ref}$ }{desired reference robot state vector at time step k }
\nomenclature[18]{$(\mathbf{x}_*)_i$}{a testing data input point that comprises $(\mathbf{x}_i, \bar{\mathbf{u}}_i, \bar{\mathbf{x}}_i, \hat{\mathbf{x}}_i)$}
\nomenclature[19]{$y_i$}{ a testing data expected output point}
\nomenclature[20]{$\mathbf{y}_*$}{the estimated residual dynamics for given $\mathbf{x}_*$ }
\nomenclature[21]{$\mathbf{z} = [\mathbf{x};\mathbf{u}]$}{ the state vector for the Gaussian Process}
\nomenclature[22]{$\delta t$ }{the integration step for Runge-Kutta 4th order method}


\printnomenclature
\section*{Introduction}

Accurate reference trajectory tracking in cluttered environments~\cite{han2023rda} is still a challenging research problem. Sensing capabilities, e.g., how far the sensor can observe the environment, and computational and power capabilities are the main constraints that face when solving such research problems. Therefore, the quadrotor's full maneuverability can not be exploited due to those constraints. On the contrary, when the problem formulation uses a simplified motion model instead of a complex dynamical model, the residual dynamical error arises between the actual controller and desired control commands.

\begin{figure}[h!]
    \centering
    \includegraphics[width=0.55\linewidth]{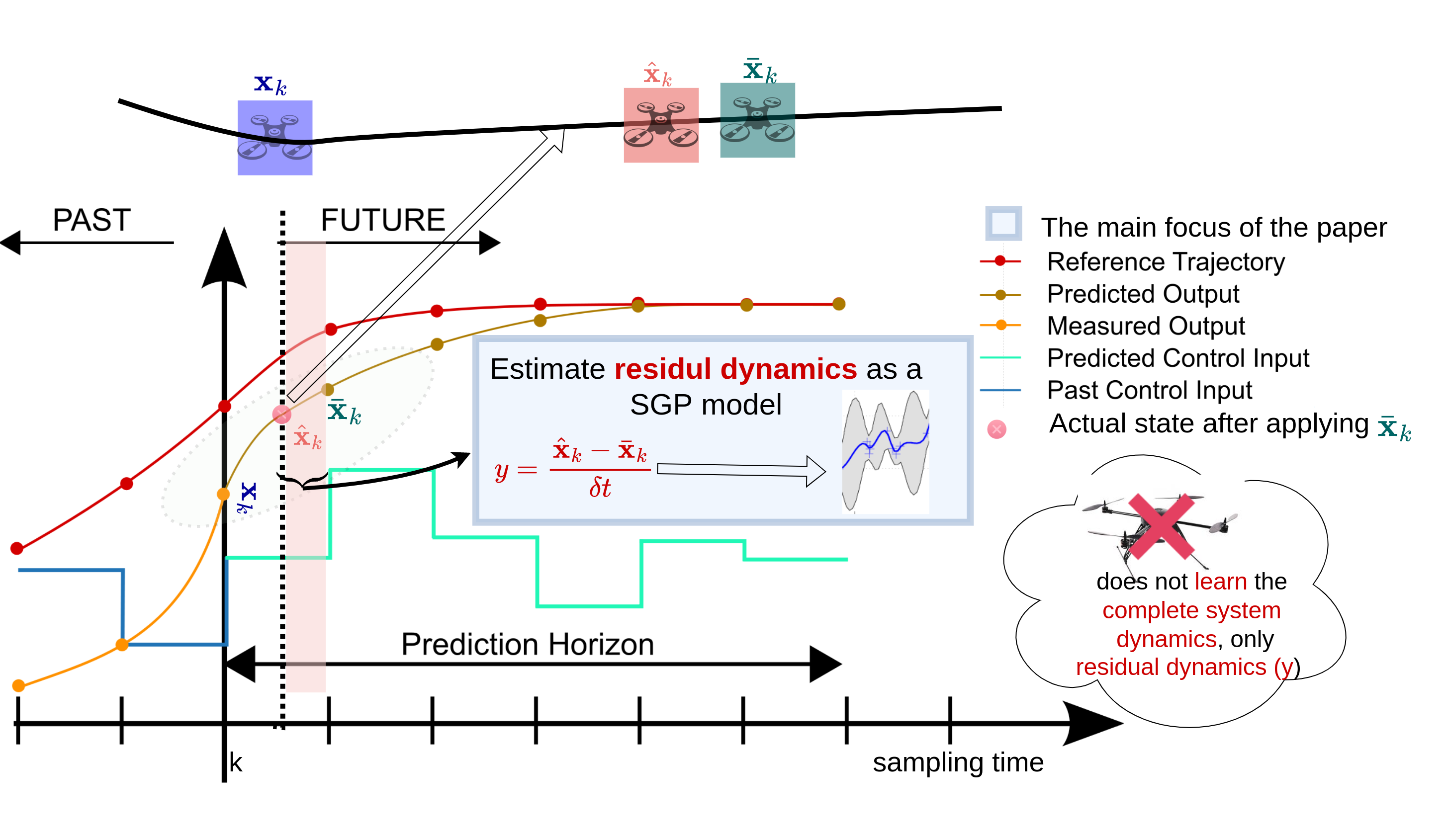}
    \caption{The high-level control command generation is based on Model Predictive Control (MPC). Residual dynamics ($y$) that arise between high-level control generated by MPC and low-level control generated by flight controller can not be estimated analytically. Hence, the Sparse Gaussian Process(SGP)-based learning technique is proposed to estimate $y$, where $\mathbf{x}_k$, $\bar{\mathbf{x}}$, and $\hat{\mathbf{x}}$ are denoted current state, next desired state, and actual state after applying the current control to the system, respectively.}
    \label{fig:trajectory_planning}
\end{figure}

For controlling a quadrotor through high-speed agile maneuvers, it is necessary to consider aerodynamic drag effects, which act as the main component of residual dynamics and external aerodynamic effects, e.g., wind, that is applied on the quadrotor in addition to other constraints: dynamics and obstacle constraints. However, several studies related to agile maneuvers ~\cite{neunert2016fast}, \cite{9422918}, \cite{rojas2021board}, \cite{song2020learning}, \cite{8424034} do not consider such effects, which are very difficult to incorporate when modelling system dynamics, except approximating the quadrotor dynamics with simplified motion model~\cite{torrente2021data}. Even if those effects are incorporated, the necessary external aerodynamic effects are difficult to obtain due to high-computational demands that leverage real-time performance. In other words, model complexity is constrained by the computational capabilities of the onboard controller. Nonetheless, such aerodynamic effects produce negligible impact for the low-speed maneuvers since dynamic effects can be neglected, considering only the kinematic modelling, especially in the plan-based control paradigm. Such a paradigm consists of two stages: planning and controlling. In the planning stage, a simplified motion model is utilized for generating near-optimal control policy, whilst in the controlling stage, the controller depends on the input from the planner and generates sufficient control commands based on the simplified motion model. However, utilizing an approximated motion model in the planning stage produces a dynamical error (residual dynamics) between the planner and the low-level controller that is hard to estimate analytically (Fig.~\ref{fig:trajectory_planning}).

In this research, we used a planner that consists of two sub-planners: a local planner that consists of a 4 degree of freedom model (simplified motion model)~\cite{kulathunga2021trajectory} and a replanner~\cite{kulathunga2022optimization} that can push the reference trajectory out of obstacle zones. The local planner is formulated as a hard constraint optimization problem, whereas the replanner is formulated as a box-constrained function minimization problem, which iteratively refines the reference trajectory by pushing away from the obstacle zones. To compensate for the limitations of the simplified motion model, a learning mechanism based on the Sparse Gaussian Process (SGP) was introduced. This learning mechanism modeled the residual dynamics ($y$)  that arose from the discrepancy between the simplified motion model and the actual system dynamics. As depicted in Fig.~\ref{fig:trajectory_planning}, the residual dynamics is determined by $y = g(\mathbf{z}), \; \mathbf{z} = [\mathbf{x}, \mathbf{u}]$, where $\mathbf{g}$ is the model that estimates residual dynamics, given current state and current control ($\mathbf{x}, \mathbf{u}$). The proposed approach employs velocity residual as the residual dynamics to integrate with the local planner's nominal dynamics, effectively reducing the discrepancy between the estimated control output from the local planner and the desired control that the simplified motion model is incapable of estimating.

\section*{Related works}

Most recent trajectory generation and tracking methods are proposed either without considering obstacles or considering partially known or simplified obstacles, e.g., gates. For example, in~\cite{romero2022time} they have achieved 60km/h on a defined racing track. This approach uses Model Predictive Contouring Control (MPCC)~\cite{romero2021model} for local replanning at such a speed. However, such methods are not capable of flying through cluttered environments due to generating a time-optimal trajectory, ensuring safety and feasibility that are limited by onboard computational capacity. Graph-based planning techniques often rely on geometric methods like Jump Point Search (JPS)~\cite{liu2017planning}, RRT~\cite{kulathunga2020real}, and A~\cite{chen2020active} to generate an initial trajectory by identifying a path or a set of waypoints for a limited horizon. This initial trajectory serves as a guide for subsequent path-planning stages. This trajectory generation can be achieved by different methods, including minimum snap~\cite{mellinger2011minimum}, B-spline~\cite{kulathunga2022optimization}. These generated trajectories, in general, do not guarantee dynamic feasibility. Kinodynamic-based trajectory planning~\cite{ding2019efficient}, \cite{kulathunga2021path} addressed these issues through kinodynamic search, whilst ensuing dynamic feasibility. Despite addressing such issues,  kinodynamic-based trajectory planning remains a highly computational footprint. In ~\cite{liu2017search}, the authors developed a kinodynamic-based framework, yet the framework failed to guarantee consistent smoothness. 

Hierarchical trajectory planning~\cite{kulathunga2022optimization}, e.g., global and local planning, is one of the ways that can address all the aforementioned issues when planning in a cluttered environment. In such planning, a global planner~\cite{singh2017robust} tries to refine the desired trajectory considering safety constraints. Along with that, a local planner~\cite{kulathunga2021trajectory} ensures the dynamic feasibility and imposes safety constraints. In the DARPA subterranean challenge~\cite{orekhov2022darpa}, Tranzatto, M. et al.~\cite{tranzatto2022cerberus} and De Petris, P. et al.\cite{de2022rmf} used a modified version of DJI M100 with position level control\cite{dji} relying on a highly accurate localisation module.  Chung, T. H. et al.\cite{chung2023into}, and Arm, P. et al.\cite{arm2023scientific} used multiple depth sensors to build highly accurate elevation maps and groundbreaking local motion planning techniques, enabling robots to navigate effectively in complex
and cluttered underground environments ensuring planning outputs are feasible for low-level controllers targeting exploration tasks. On the other hand, with fewer sensing capabilities, incorporating fully-fledged system dynamics makes it harder to solve trajectory planning whilst avoiding obstacles in real-time. Hence, the approaches that employ approximated system dynamics use external and internal disturbances alongside approximated dynamics~\cite{singh2017robust}, \cite{wang2022kinojgm}, \cite{guerrero2013quad}, \cite{mehndiratta2020gaussian}. Also, recent advancements in adaptive robust nonlinear control-based techniques~\cite{flores2022robust, nian20232} have demonstrated promising results for controlling aerial vehicles. Most of these techniques can be categorized under different variations of receding horizon-based control. A special variant of such control, GP-MPC (Gaussian Process MPC) excels in estimating unknown quantities, such as residual dynamics, internal disturbances, and external disturbances. A recent line of work~\cite{torrente2021data}, \cite{hewing2019cautious}, \cite{kabzan2019learning}, \cite{cao2017gaussian} has been investigating how to incorporate Gaussian Process (GP) to learn system dynamics entirely or partially considering external disturbances, for example, wind~\cite{mehndiratta2020gaussian}, from data, targeting applications such as reference trajectory tracking and reaching to a specified goal. The choice of GPs is preferred for system identification-related tasks, e.g., residual dynamics modelling, compared to more expressive modelling techniques such as neural networks in the recent past. GPs are preferred due to the modelling constraints for receding horizon planning, e.g., MPC and NMPC. It is sophisticated to develop the back-propagation algorithm~\cite{cilimkovic2015neural} to estimate gradients of the cost function when formulating receding horizon planning problems within the optimization solver. However, for GPs, it is relatively easy to implement gradient estimation. Hence, variants of GPs are proposed for system identification-related tasks more often than the other techniques.

Instead of learning a complete dynamical model, the authors of~\cite{desaraju2018leveraging} proposed to learn only time-varies state uncertainty using Gaussian belief propagation. Consequently, learning only the sub-portion of time-varies state or/and control uncertainty as the residual dynamics yields quite promising results~\cite {torrente2021data}. The latter training technique requires less data to learn the residual dynamics. Yet it is advisable to have more training data to improve the robustness to learn the residual dynamics. However, a large amount of training data is problematic when GP is applied for real-time applications. There are several ways to reduce the overhead that arises when utilizing a high volume of training data~\cite{quinonero2005unifying}: select a set of points, i.e, inducing points~\cite{wilson2015thoughts}, \cite{rasmussen2010gaussian}, that approximate the original training data distribution; exploit the structure of the formation of GP~\cite{wilson2015kernel}, \cite{cunningham2008fast}, \cite{flaxman2015fast}, e.g., Kronecker Structure~\cite{werner2008estimation}; and use variational methods. Sparse GP~\cite{titsias2009variational} is such a variant that can be used in different ways to obtain inducing points: Subset of Regressors (SoR), Deterministic Training Conditional (DTC), Fully Independent Training Conditional (FITC)~\cite{rasmussen2010gaussian}, and Structured Kernel Interpolation (SKI)~\cite{wilson2015kernel}. The complexity reduces from $O(n^3)$ to $O(nm^3)$, where n and m are the numbers of points ($n >m$) that are used to model GP and Sparse GP, respectively.  Inducing point methods are quite appropriate for the real-time setting. However, model accuracy degrades as $m$ decreases, which can be seen as a trade-off between model expressiveness reduction versus computational complexity. 

\begin{table}[h!]
\centering
\caption{Comparative analysis of Gaussian Process-based residual dynamics learning techniques with the proposed technique}
\label{t:comparision1}
\begin{tabular}{|l|l|l|l|l|l|}
\hline
Approach                                                         & \begin{tabular}[c]{@{}l@{}}Quadrotor model\end{tabular} & Type of learning                                                                & \begin{tabular}[c]{@{}l@{}}Obstacle\\ avoidance\end{tabular} & \begin{tabular}[c]{@{}l@{}}Experiments\\ (real-world or  simulated)\end{tabular} & Speed-profile                                                         \\ \hline
 
\begin{tabular}[c]{@{}l@{}}KNODE-MPC  \cite{chee2022knode} \end{tabular} & KNODE                                                     & \begin{tabular}[c]{@{}l@{}}residual \\ dynamics\end{tabular}                    & no                                                           & \begin{tabular}[c]{@{}l@{}}real-world (off-board \\ computation)\end{tabular}    & \begin{tabular}[c]{@{}l@{}}low-speed\\  (\textless 1m/s)\end{tabular} \\ \hline
\begin{tabular}[c]{@{}l@{}}Data-Driven MPC  \cite{torrente2021data} \end{tabular}                                                   & simplified                                                & \begin{tabular}[c]{@{}l@{}}aerodynamic\\ effects\end{tabular}                   & no                                                           & \begin{tabular}[c]{@{}l@{}}real-world (on-board \\ computation)\end{tabular}     & \begin{tabular}[c]{@{}l@{}}high-speed \\ (up to 14m/s)\end{tabular}   \\ \hline
\begin{tabular}[c]{@{}l@{}}KinoJGM  \cite{wang2022kinojgm}  \end{tabular}                                                       & full-state                                                & \begin{tabular}[c]{@{}l@{}}unpredictable \\ aerodynamic  effects\end{tabular} & yes                                                          & only simulated                                                                     & \begin{tabular}[c]{@{}l@{}}low-speed \\ (\textless 3m/s)\end{tabular} \\ \hline
\begin{tabular}[c]{@{}l@{}}The proposed \\ technique\end{tabular} & simplified                                                & \begin{tabular}[c]{@{}l@{}}residual \\ dynamics\end{tabular}                    & yes                                                          & \begin{tabular}[c]{@{}l@{}}real-world (on-board \\ computation)\end{tabular}     & \begin{tabular}[c]{@{}l@{}}low-speed \\ (\textless 3m/s)\end{tabular} \\ \hline
\end{tabular}
\begin{tablenotes}
     \item[1]  KNODE: Knowledge-based Neural Ordinary Differential Equations to augment a model.
\end{tablenotes}
\end{table}

In this work, we present how to reduce residual dynamics that arise between high-level planning and low-level controlling by considering reference trajectory tracking as an example in the plan-based control paradigm (Table.\ref{t:comparision1}). The proposed approach is not limited to the considered example. It is valid for any motion planner which depends on a low-level controller. Depending on the representation of the motion model, the residual dynamics are defined by selecting a set of states and controlling input appropriately. The uniqueness of the proposed approach is that it is invariant to the geometry representation of the trajectory. Hence, once the residual dynamics model is trained, it can be deployed without retraining. The following sections explain the training process of such residual dynamics model and deploy it in hardware, which shows the effectiveness of incorporating residual dynamics modelling in high-level planning. 

Our \textbf{contributions} are as follows:
\begin{enumerate}
    \item To address the limitations of traditional motion models in capturing the actual system dynamics of aerial vehicles, we propose a hybrid approach that combines a simplified motion model for high-level planning with a learning-based technique, Sparse Gaussian Process, to estimate residual dynamics. This approach enables real-time performance by running the entire algorithm on an onboard computer.

    \item To augment the replanner's~\cite{kulathunga2022optimization} capability to circumvent local minima, we introduce a mechanism that dynamically regenerates the reference trajectory whenever the quadrotor's actual position deviates substantially from the initial reference trajectory. This enhancement effectively precludes the quadrotor from becoming ensnared in local minima during the receding horizon control process.
    
\end{enumerate}

\section*{Methodology}


This paper tackles the challenge of using a less accurate model (kinematic or dynamic) with Model Predictive Control (NMPC) by learning the residual errors between that approximated model and the actual model. Instead of a full 6-degree-of-freedom dynamic model, we employ a 4-degree-of-freedom model and propose a Sparse Gaussian Process technique to reduce the impact of dynamic factors that depend on the operating conditions and can hardly be modelled analytically in advance. Crucially, we aim to maintain trajectory planner performance, so we focus on learning the rate of velocity changes as residual dynamics. This choice is invariant to the trajectory's geometric representation, allowing offline learning of the residual dynamics distribution. By generating diverse trajectories, we capture the latent distribution of residual dynamics. We then augment the nominal model with this learned residual dynamics model and demonstrate it in online NMPC  as a feedback controller, especially for the quadrotor in the plan-based robot control paradigm.

The plan-based robot control paradigm comprises two stages: planning and controlling. The planning stage can be a combination of several planners that run simultaneously or cascade planners~\cite{xu2020backstepping} that run sequentially. In this work, the planning stage is formed by two planners: a replanner and a local planner. We have used the approach proposed in~\cite{kulathunga2022optimization} as our replanner. It executes continuous refinement of the provided reference trajectory out of obstacle regions. The proposed learning-based residual dynamics are augmented into a local planner~\cite{kulathunga2021trajectory}. Further, the subsections are organized as follows: formulating the local planner with residual dynamics in the Residual Augmented Quadrotor Motion Model; laying down training strategies of residual dynamics in the Nonlinear Model Predictive Control local planner, and estimating approximated distribution to represent the residual dynamics in Sparse GP Regression. 


\subsection*{Residual Augmented Quadrotor Motion Model}
The proposed residual dynamics learning framework can be integrated with various quadrotor kinematical and dynamical models when formulating NMPC. While the choice of model can significantly impact the effectiveness of the approach, it also introduces various challenges, including high computational demands, being stuck in local minima, low convergence rate, and real-time applicability. In NMPC, finding a delicate balance between avoiding local minima and managing computational costs is crucial. A longer prediction horizon helps evade local minima, but it also increases the number of optimization variables, demanding more computational resources. Several strategies exist to reduce these variables, but each has its own drawbacks. Increasing the discretization interval ($\delta t$): This relies heavily on the accuracy of the motion model. If the model is approximated, it can lead to unrealistic control commands. Increasing maximum control inputs (velocity, acceleration): this raises the risk of collisions with obstacles. Using high-degree-of-freedom models: if some degrees of freedom aren't fully utilized, unnecessary computational overhead is introduced. In this work, we chose a 4-degree-of-freedom model ($\mathbf{f}_{norm}(\mathbf{x}, \mathbf{u})$)\cite{kulathunga2021trajectory}, aligning with our focus on low-speed maneuvering. To compensate for potential inaccuracies, we employed Sparse Gaussian Regression to model residual dynamics. This approach strikes a balance between avoiding local minima and maintaining computational efficiency, demonstrating the intricate link between these two factors in NMPC. Let $\mathbf{f}_{est}(\mathbf{x}, \mathbf{u})$ be the augmented quadrotor motion model:

\begin{equation}\label{eq:estimated_model}
    \dot{\mathbf{x}} = \mathbf{f}_{est}(\mathbf{x}, \mathbf{u}) =  \mathbf{f}_{norm}(\mathbf{x}, \mathbf{u}) +  B_z g(\mathbf{z}),
\end{equation} where the state vector and the control inputs, defined $\mathbf{x}$ and $\mathbf{u}$, respectively. Depending on which states and controls are learnt, $B_z$ gives the appropriate transformation, defined in the following sections. The Gaussian process was used to model the residual dynamics $g(\mathbf{z}) \sim N(\mu, \Sigma_p), \mathbf{z} = [\mathbf{x};\mathbf{u}]$, where $\Sigma_p$ is a i.i.d (independent and identically distributed) process noise and $\mu$ estimated mean value. The training data were centered, i.e., zero mean, before stating the learning process, which consists of only the diagonal covariance (or assuming no correlation between the two states of residual dynamics). The explicit Runge-Kutta 4th order algorithm was used to incorporate $\mathbf{f}_{est}(\mathbf{x}, \mathbf{u})$ motion model in the discrete-time $\mathbf{f}_d(\mathbf{x}_k, \mathbf{u}_k, \delta t)$ with integration step size $\delta t$ as $\mathbf{x}_{k+1} = \mathbf{f}_d(\mathbf{x}_k, \mathbf{u}_k, \delta t),$  where k depicts the time index. Since we rely on a less accurate model (or an approximate model) using the Runge-Kutta 4th order method\cite{westermann2024accuracy} compared to the Eular, reduce the integration error of the discrete system dynamics.

\begin{figure*}[h!]
    \centering
    \includegraphics[width=0.8\linewidth]{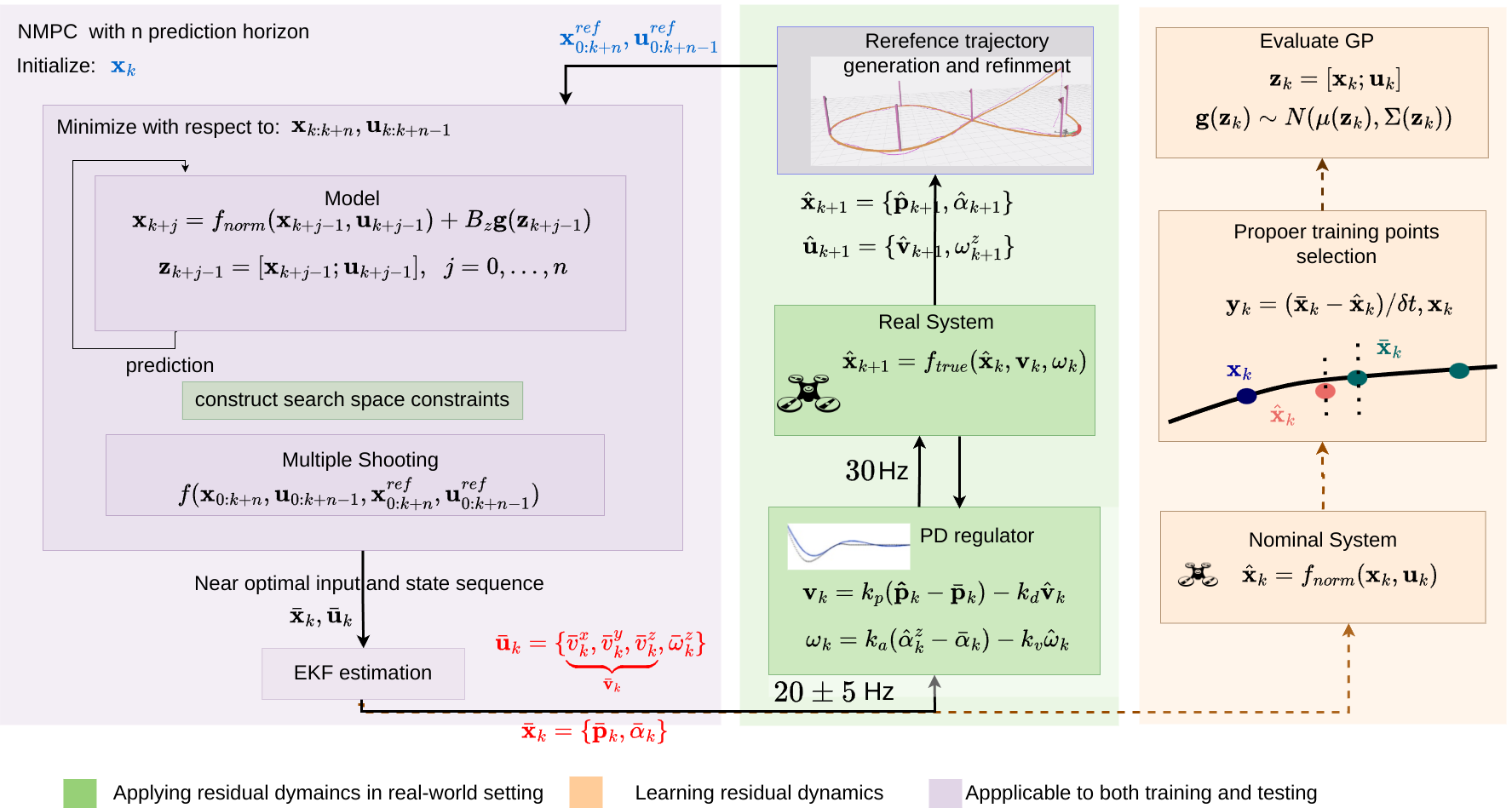}
    \caption{The high-level overview of the proposed framework for learning the residual dynamics. Design matrix $B_{\mathbf{z}}$ defines which states and control inputs should be trained. Learned residual dynamics  $g(\mathbf{z}_{k+i-1})$ is added to nominal dynamics $f_{norm}(\mathbf{x}_{k+i-1}, \mathbf{u}_{k+i-1})$ when formulating NMPC by using the multiple shooting technique. In the training stage, collect the data $D = \{X, \mathbf{y}\} = \{(\mathbf{x}_i, \bar{\mathbf{u}}_i, \bar{\mathbf{x}}_i, \hat{\mathbf{x}}_i), y_i\}, i=0,...,n$ for n number of times, where $(\mathbf{x}_i, \bar{\mathbf{u}}_i, \bar{\mathbf{x}}_i$, and $\hat{\mathbf{x}}_i)$ denote current state, estimated near-optimal control inputs and state after applying NMPC, and actual system state after applying estimated control inputs, respectively. We used the same PD regulator that was proposed in~\cite{kulathunga2021trajectory}. This work focuses on the DJI M100 quadrotor as the representative real system.}
    \label{fig:system_overview}
\end{figure*}

\subsection*{Nonlinear Model Predictive Control Based Local Planner}

To obtain augmented residual motion model $\mathbf{f}_{d}(\mathbf{x}, \mathbf{u}, \delta t)$, the local planner calculates near-optimal control policy. However, the local planner does not guarantee optimality because NMPC-based planners calculate approximated control policy heuristically. In complex and cluttered environments, generating a convex representation of free space is generally challenging. However, the incremental Euclidean Distance Transformation Map (EDTM)~\cite{kulathunga2021trajectory}, when applied along the reference trajectory, can provide accurate obstacle poses. These poses can be incorporated as hard constraints in the NMPC formulation, effectively addressing the non-linearity of obstacle constraints. NMPC is preferred over linear or convex MPC due to its ability to generate feasible solutions even when dealing with non-linear constraints. However, NMPC is susceptible to getting stuck in local minima, unlike convex MPC which guarantees global optimality. The following section outlines a proposed method to mitigate this issue.

For a specified time index k, a corresponding slice of the reference trajectory $[\mathbf{x}_{k}^{ref},..., \mathbf{x}_{k+N}^{ref}, \mathbf{u}_{k}^{ref},..., \mathbf{u}_{k+N}^{ref}]$ and the current state $\textbf{x}_{k}$, and close-in obstacles $g_2(\mathbf{w})$ information are available. The local planner, which was designed using multiple shooting technique, generates near-optimal control policy $\mathbf{w} = [\bar{\textbf{u}}_{k}, \hdots , \bar{\textbf{u}}_{k+N-1}, \bar{\textbf{x}}_{k}, \hdots, \bar{\textbf{x}}_{k+N}]$ as follows: 

\begin{equation}\label{eq:nmpc}
\begin{aligned}
 \min_{\mathbf{w}} & \quad  J(\mathbf{w}) =  \sum_{i=0}^{N}{\left \| \bar{\mathbf{x}}_{k+i} -\mathbf{x}^{ref}_{k+i} \right \|_Q^2 + \left \| \bar{\mathbf{u}}_{k+i} -\mathbf{u}^{ref}_{k+i} \right \|_R^2} \quad \\
 \textrm{s.t.}  &  \quad   g_1(\mathbf{w}) = 0 , \; \; g_2(\mathbf{w}) \leq 0 , \\
   &  \quad \mathbf{x}_{min} \leq \bar{\mathbf{x}}_{k+i} \leq \mathbf{x}_{max} \quad \forall 0 \leq i \leq N , \\ 
  &   \quad -\mathbf{v}_{max} \leq \bar{\mathbf{u}}_{k+i} \leq \mathbf{v}_{max} \quad \forall  0 \leq i \leq N-1 ,
\end{aligned}
\end{equation} where the performance index (or objective function) $J(\mathbf{w})$ aims to minimize the proportional error $\bar{\mathbf{x}}_{k+i} -\mathbf{x}^{ref}_{k+i}$ and the control effect $\bar{\mathbf{u}}_{k+i} -\mathbf{u}^{ref}_{k+i}$ by employing a quadratic cost function. The continuous and differentiable nature of quadratic functions facilitates smooth optimization, enabling efficient solution via sequential quadratic programming (SQP)~\cite{boggs1995sequential}. The prediction horizon length of NMPC, denoted N. Both maximum and minimum pairs of velocities and positions, denoted $\mathbf{v}_{max}, \mathbf{x}_{max}$ and $\mathbf{v}_{min}, \mathbf{x}_{min}$, respectively. Both the $Q$ and $R$ matrices are positive semi-definite and positive definite, respectively.  These matrices govern the importance of state variables and control inputs in the cost function, $J(\mathbf{w})$.
 To obtain the discrete dynamical model, the Runge-Kutta 4th order algorithm is applied to the quadrotor's motion dynamics, resulting in the discrete dynamical model $\mathbf{x}_{k+1} = \mathbf{f}_d(\mathbf{x}_k, \mathbf{u}_k, \delta t)$, where $\delta t$ is the time step. The system dynamic constraints are formulated based on the given discrete dynamical model $\mathbf{f}_d$(eq.2)~\cite{kulathunga2021trajectory}.

\begin{equation}
\begin{aligned}
    \quad  & g_1(\mathbf{w}) = \begin{bmatrix}
 \textbf{x}_{k}- \bar{\textbf{x}}_{k+i} \\
   \vdots \\
 \mathbf{f}_d(\bar{\textbf{x}}_{k+N-1}, \bar{\textbf{u}}_{k+N-1}, \delta t)-\bar{\textbf{x}}_{k+N}
 \end{bmatrix}.
\end{aligned}
\end{equation} At each time step k, the desired reference pose and control are denoted by $\mathbf{x}^{ref}_{k}$  and $\mathbf{u}^{ref}_{k}$, respectively. The reference trajectory, which provides the desired poses and control inputs, can be represented as a $d$th-degree polynomial function of time. Consequently, desired derivatives up to order $d-1$ can be obtained at each time step $k$. Cubic uniform B-spline was employed to define the initial reference trajectory, which was subsequently refined using the global planner~\cite{kulathunga2022optimization}. The obstacle constraints were formulated as a set of nonlinear hard constraints, $g_2(\mathbf{w})$, as follows:
\begin{equation}
\begin{aligned}
    \quad g_2(\mathbf{w}) =  \begin{bmatrix}
 dis(\textbf{x}_j^{o}, \bar{\textbf{x}}_{k}) \\
 \vdots \\
 dis(\textbf{x}_j^{o}, \bar{\textbf{x}}_{k+N}) \\
 \end{bmatrix}, \; j = 1,...,N_{o},i = 0,...,N,
\end{aligned}
\end{equation}  where $N_{o}$ is the number of obstacles at time k, and $dis(\textbf{x}_j^{o}, \bar{\textbf{x}}_{k+h})$ is calculated as $-\sqrt{(x_j^{o} - x_{k+h})^2 + (y_j^{o}-y_{k+h})^2 + (z_j^{o}-z_{k+h})^2} + d^{o},$  
where $d^{o}$ is the safe zone distance between a quadrotor and close-in obstacles, denoted by $\textbf{x}_j^{o} \in \mathbb{R}^3, j=1,..,N_o$. For the descriptive formulations of $ g_1(\mathbf{w})$, $ g_2(\mathbf{w})$, $\mathbf{x}^{ref}_{k}$, and $\mathbf{u}^{ref}_{k}$, refers to~\cite{kulathunga2021trajectory, kulathunga2022optimization}.

\subsection*{Augmented Residual Dynamics Learning with Gaussian Process(GP)}
To learn residual dynamics $g(\mathbf{z})$ eq.(\ref{eq:estimated_model}), the first step is to collect a training dataset. As shown in Fig.~\ref{fig:system_overview}, generate a set of random reference trajectories and formulate the local planner eq.(\ref{eq:nmpc}) considering only $\mathbf{f}_{norm}(\mathbf{x}, \mathbf{u})$. Let training dataset D be $\{X, \mathbf{y}\}$, where $ X = \{(\mathbf{x}_i, \bar{\mathbf{u}}_i, \bar{\mathbf{x}}_i, \hat{\mathbf{x}}_i)\}_{i=0}^n$ and $\mathbf{y} = \{y_i\}_{i=0}^n, i=0,...,n$. Terms $\mathbf{x}_i, \bar{\mathbf{u}}_i, \bar{\mathbf{x}}_i$ and $\hat{\mathbf{x}}_i$ are current state, (predicted) control input and state using the local planner, and actual states after applying $\bar{\mathbf{u}}_i$ to the system, respectively, at time $t_i$. Residual dynamics $y_i$ at $t_i$ is determined by
\begin{flalign}\label{eq:gaussian_model_}
    \begin{aligned}
        y_i = \frac{\hat{\mathbf{x}}_i - \bar{\mathbf{x}}_i}{\delta t_i} = g(\mathbf{z}_i), \; \mathbf{z}_i = [\mathbf{x}_i, \mathbf{u}_i]
    \end{aligned}
\end{flalign}Thus, given a training dataset D, Gaussian Process is employed to learn $g(\mathbf{z}_i)$. Gaussian Process, in general, can be used to infer a distribution (prior) over function directly. Hence, residual dynamics can be determined as follows: 
\begin{flalign}
\label{eq:gau_linear}
    y_i = g({\mathbf{z}_i}) + \epsilon_i, \; \; \epsilon_i \sim N(0, \sigma_n),
\end{flalign} where $\epsilon_i$ is the corresponding n white noise realization $\sigma_n$. Given an unseen $\mathbf{x}_*$ data points ( or test dataset), GP, which is a way to estimate the posterior predictive distribution, estimates corresponding residual dynamics $\mathbf{y}_*$, that can be formulated as follows:

\begin{flalign}\label{eq:gau_posrerior_predictive}
    p(\mathbf{y}_*|\mathbf{x}_*,X, \mathbf{y}) = \int_{\mathbf{s}}p(\mathbf{y}_*,\mathbf{s}|X, \mathbf{y}, \mathbf{x}_*) d\mathbf{s}  = \int_{\mathbf{s}} p(\mathbf{y}_*|\mathbf{x}_*, \mathbf{s})p(\mathbf{s}|X,\mathbf{y})d\mathbf{s} 
     \sim N\Bigl(\frac{1}{\sigma_n^2}\mathbf{x}_*^{\top}A^{-1}X\mathbf{y}, \mathbf{x}_*^{\top}A^{-1}\mathbf{x}_*\Bigl),
\end{flalign} where $A = \sigma_n^{-2}XX^{\top} +\Sigma_p^{-1}$ and term $\mathbf{s}$, denoted the prior distribution of the residual dynamics. Nonetheless, linear models, e.g.,~eq.(\ref{eq:gau_linear}), suffer severely from limited expressiveness. Various methods can be used to transform input space ($\mathbf{x}_i$) into different feature spaces using basis functions; however, the model ($\mathbf{s}$) is still linear in parameters, which is analytically tractable. Let $\Phi(*)$ be such a basic function, where $*$ denotes input space. Along with that, posterior predictive distribution eq.(~\ref{eq:gau_posrerior_predictive}) becomes:
$p(\mathbf{y}_*|\mathbf{x}_*,X,\mathbf{y}) \sim N\Bigl(\frac{1}{\sigma_n^2}\Phi(\mathbf{x_*})^{\top}A^{-1}\Phi\mathbf{y}, \Phi(\mathbf{x_*})^{\top}A^{-1}\Phi(\mathbf{x}_*)\Bigl),$ where $\Phi = \Phi(X), A = \sigma_n^{-2}\Phi\Phi^{\top} + \Sigma_p^{-1}$. This posterior predictive distribution can be reformulated as  $p(\mathbf{y}_*|\mathbf{x}_*, X,\mathbf{y}) \sim N\Bigl(\sigma_*^{\top}\Sigma_p\Phi(K+\sigma_n^2I)^{-1}\mathbf{y}, \Phi_*^{\top}\Sigma_p\Phi_*  -\Phi_*^{\top}\Sigma_p\Phi(K+\sigma_n^2I)^{-1}\Phi^{\top}\Sigma_p\Phi_{*}\Bigl), $ where $\Phi(\mathbf{x}_*) = \Phi_*$ and $K = \Phi^{\top}\Sigma_p\Phi$. Feature space ($\Phi(*)$) always enters in the form of $\Phi^{\top}\Sigma_p\Phi, \Phi_*^{\top}\Sigma_p\Phi,$ or $\Phi_*^{\top}\Sigma_p\Phi_*$. Thus, the entries of these matrices have invariable form $\Phi(\mathbf{x}_{\mu 1})^{\top}\Sigma_p\Phi(\mathbf{x}_{\mu 2})$, where $\mathbf{x}_{\mu 1}$ and $\mathbf{x}_{\mu 2}$ come either from training or testing sets. Let $k(\mathbf{x}_{\mu 1}, \mathbf{x}_{\mu 2}) = \Phi(\mathbf{x}_{\mu 1})^{\top}\Sigma_p\Phi(\mathbf{x}_{\mu 2})$. Since $\Sigma_p$ is positive definite, we can define $\Sigma_p^{1/2}$. Moreover, $\Sigma_{p}^{1/2} = UD^{1/2}U^{\top}$ (using Singular Value Decomposition (SVD)). Thus,  $k(\mathbf{x}_{\mu 1}, \mathbf{x}_{\mu 2}) = \psi(\mathbf{x}_{\mu 1})\cdot \psi(\mathbf{x}_{\mu 2})$, where $\psi(\mathbf{x}_{\mu 1}) = \Sigma_p^{1/2}\Phi(\mathbf{x}_{\mu 1})$. If $\mathbf{x}_{\mu 1}$ and $\mathbf{x}_{\mu 2}$ are similar, kernel $k(\mathbf{x}_{\mu 1}, \mathbf{x}_{\mu 2})$ at these points, i.e., $f(\mathbf{x}_{\mu 1})$ and $f(\mathbf{x}_{\mu 2})$, must have similar values. 

Since data centring is, in general, applied before applying regression, a GP is completely specified by its co-variance function $k(\mathbf{x}_{\mu 1}, \mathbf{x}_{\mu 2})$ for a real process $g(\mathbf{x})$ (or $g(\mathbf{z})$, where $\mathbf{z} = [\mathbf{x};\mathbf{u}]$). We have used only $\mathbf{x}$ as the input for GP eq.(\ref{eq:gaussian_model_}). Hence, we neglected $\mathbf{u}$ from the formulation). The $g(\mathbf{x}_*) \sim GP( m(\mathbf{x}), cov(\mathbf{x}, \mathbf{x}_*))$, where terms $m(\mathbf{x}) = \mathbb{E}[g(\mathbf{x})], m(\mathbf{x_*})$, $cov(\mathbf{x}, \mathbf{x}_*) \\\nonumber = \mathbb{E}[(g(\mathbf{x})- m(\mathbf{x}))(g(\mathbf{x}_*)- m(\mathbf{x}_*))] $, denoted mean for training data, mean for testing data and covariance between training and testing data, respectively. Since it is impossible to estimate $k(\mathbf{x}_{\mu 1}, \mathbf{x}_{\mu 2})$ analytically, an approximated kernel function k, e.g., square exponential (SE)~\cite{seeger2004gaussian}, is required to estimate $\Sigma_p$ is defined as $k(\mathbf{x}_{\mu 1}, \mathbf{x}_{\mu 2}) =  \sigma_f^2 e^{(-\frac{1}{2l^2}\left | \mathbf{x}_{\mu 1} - \mathbf{x}_{\mu 2} \right |^2)} + \sigma_n^2I$, where $\sigma_f$ and $l$ are hyper-parameters that have to learn; they are unique for the selected kernel function. Along with that, GP is formed to estimate $\mathbf{y}_*$ analytically given $\mathbf{x_*}$ as follows:
\begin{flalign}\label{eq:gaussian_process}
    \begin{vmatrix}
\mathbf{y}\\ 
\mathbf{y}_*
\end{vmatrix} \sim  N \left( 0, \begin{bmatrix}
k(\mathbf{x}+\sigma_n^2I, \mathbf{x}) & k(\mathbf{x}, \mathbf{x_*})\\ 
k(\mathbf{x}_*, \mathbf{x}) & k(\mathbf{x}_*, \mathbf{x}_*)
\end{bmatrix}\right)
\end{flalign} Thus, using eq.(\ref{eq:gaussian_process}), the mean and covariance of the testing dataset can be fully determined as $m(\mathbf{x_*}) = k(\mathbf{x}_*, \mathbf{x})[k(\mathbf{x}, \mathbf{x})+\sigma_n^2I]^{-1}\mathbf{y}$ and 
 $cov(\mathbf{x}_*, \mathbf{x}_*) = k(\mathbf{x}_*, \mathbf{x}_*)-k(\mathbf{x}_*, \mathbf{x})[\mathbf{x}, \mathbf{x}+\sigma_n^2I]^{-1}k(\mathbf{x}, \mathbf{x}_*)$, respectively. 

\subsection*{Sparse GP Regression}
Let $\mathbf{y}_i = \mathbf{g}_i + \epsilon_i, \mathbf{g}_i = \mathbf{g}(\mathbf{x}_i)$, where $\mathbf{g}= \{\mathbf{g}_i\}_{i=1}^n$ are the latent function values. Once GP is estimated, the prior $p(\mathbf{g})$ can be determined. Therefore, the joint probability model $p(\mathbf{y}, \mathbf{g}) = p(\mathbf{y}|\mathbf{g})p(\mathbf{g})$ can be estimated. Sparse GP Regression aims to minimize the distance between the exact Gaussian distribution and a proposed posterior Gaussian distribution using variational approximation in which the number of training samples (m) is less than the number of training samples (n) for exact GP .  Let $p(\mathbf{h}|\mathbf{y})= \int p(\mathbf{h}|\mathbf{g})p(\mathbf{g}|\mathbf{y})d\mathbf{g}$ be such a posterior distribution, where $p(\mathbf{h}|\mathbf{g})$ denotes the prior conditional probability on the selected a set of auxiliary inducing points $\mathbf{x}_m$ and $\mathbf{h} = \mathbf{g}_j +\epsilon_j, \epsilon_j \sim N(0, \sigma_m), j= \{0,...,m\}$. $\epsilon_j$ is the corresponding m white noise realization $\sigma_m$.  After selecting such points,  $\mathbf{g}_m$ be the approximated GP model. Therefore, $p(\mathbf{h}|\mathbf{y})$ can be reformulated as: $p(\mathbf{h}|\mathbf{y}) = \int p(\mathbf{h}|\mathbf{g}_m,\mathbf{g})p(\mathbf{g}|\mathbf{g}_m, \mathbf{y})p(\mathbf{g}_m|\mathbf{y})d\mathbf{g}d\mathbf{g}_m$. Suppose $\mathbf{h}$ and $\mathbf{g}$ are conditionally independent given $\mathbf{g}_m$, i.e., $p(\mathbf{h}|\mathbf{g}_m,\mathbf{g}) = p(\mathbf{h}|\mathbf{g}_m)$ or $\mathbf{g}_m$ is sufficient to describe the distribution $\mathbf{g}$:   
\begin{flalign}
        q(\mathbf{h}) = p(\mathbf{h}|\mathbf{y}) = \int p(\mathbf{h}|\mathbf{g}_m)p(\mathbf{g}|\mathbf{g}_m)\Phi(\mathbf{g}_m)d\mathbf{g}d\mathbf{g}_m  = \int p(\mathbf{h}|\mathbf{g}_m)\Phi(\mathbf{g}_m)d\mathbf{g}_m = \int q(\mathbf{h},\mathbf{g}_m)d\mathbf{g}_m
\end{flalign} where $\Phi(\mathbf{g}_m) = p(\mathbf{g}_m|\mathbf{y}) \sim N(\mathbf{\mu}, A)$. With that, mean and covariance of the approximated posterior GP are determined as 
\begin{equation*}
    \begin{aligned}
         m(\mathbf{x}_*) = k(\mathbf{x}_*,\mathbf{x}_m)k(\mathbf{x}_m,\mathbf{x}_m)^{-1}\mathbf{\mu}, \\
         cov(\mathbf{x}_*, \mathbf{x}_*) = k(\mathbf{x}_*, \mathbf{x}_*) - k(\mathbf{x}_*\mathbf{x}_m)k(\mathbf{x}_m,\mathbf{x}_m)^{-1}k(\mathbf{x}_m,\mathbf{x}_*) + k(\mathbf{x}_*,\mathbf{x}_m)k(\mathbf{x}_m,\mathbf{x}_m)^{-1}Ak(\mathbf{x}_m,\mathbf{x}_m)^{-1}k(\mathbf{x}_m,\mathbf{x}_*)
    \end{aligned}
\end{equation*}respectively. Inducing points $x_m$ are estimated to reduce the distance between approximated and actual Gaussian distributions. The Kullback-Leibler ($ \mathbb{KL}$) divergence can be used to minimize the distance between the approximated posterior $q(\mathbf{h})$ and the exact posterior $p(\mathbf{g}|\mathbf{y})$. The optimal estimation $q^*(\mathbf{h}) =  \argmin_{\mathbf{\Phi(\mathbf{g}_m)}} \mathbb{KL}[q\Bigl(\mathbf{h}; \Phi(\mathbf{g}_m)\Bigl)||p(\mathbf{h}|\mathbf{y})]$. Using Bayesian inference, i.e., $p(h|y)=p(h,y)/p(y)$, $ \mathbb{KL}$ divergence can be expanded in the following way:

\begin{flalign}\label{eq:kl_divergance}
        \mathbb{KL}[q\Bigl(\mathbf{h}; \Phi(\mathbf{g}_m)\Bigl)||p(\mathbf{h}|\mathbf{y})] = \mathbb{E}[log\Bigl(q(\mathbf{h};\Phi(\mathbf{g}_m))\Bigl)]  - \mathbb{E}[log\Bigl(p(\mathbf{y},\mathbf{h})\Bigl)] +  \mathbb{E}[log\Bigl(p(\mathbf{y})\Bigl)]
\end{flalign} However, $p(\mathbf{y})$ is an intractable integral, yet  $p(\mathbf{y})$ is independent of $q(\mathbf{h}; \Phi(\mathbf{g}_m))$. Thus, by maximizing the first two terms (evidence lower bound $ELBO(q) = L(\mu, A, \mathbf{x}_m)$):
\begin{flalign}\label{eq:elbo1}
     ELBO(q) = \mathbb{E}[log\Bigl(p(\mathbf{y},\mathbf{h})\Bigl)] - \mathbb{E}[log\Bigl(q(\mathbf{h};\Phi(\mathbf{g}_m))\Bigl)]
\end{flalign} true log marginal likelihood $log(p(\mathbf{y}))$ can be calculated. Thus, relationship between $ELBO(q)$ and $\mathbb{KL}[q(\mathbf{h}; \Phi(\mathbf{g}_m))||p(\mathbf{h}|\mathbf{y})]$ can be determined considering eq.(\ref{eq:kl_divergance}) and eq.(\ref{eq:elbo1}) as $
     ELBO(q) =  \mathbb{E}[log\Bigl(p(\mathbf{y}|\mathbf{h})\Bigl)] - \mathbb{KL}[q\Bigl(\mathbf{h}; \Phi(\mathbf{g}_m)\Bigl)||p(\mathbf{h})]$. Similarly, log marginal likelihood $log(p(\mathbf{y}))$ can also be determined by
$log(p(\mathbf{y})) = ELBO(q) + \mathbb{KL}[q\Bigl(\mathbf{h}; \Phi(\mathbf{g}_m)\Bigl)||p(\mathbf{h}|\mathbf{y})]$. Since $\mathbb{KL}[q(\mathbf{h}; \Phi(\mathbf{g}_m))||p(\mathbf{h}|\mathbf{y})] \geq 0$ for any true distribution, i.e. $\mathbb{KL}$ is a distance, $log (p(\mathbf{y})) \geq ELBO(q)$ should hold. Thus, this inequality maximizes the $ELBO$ on $\mathbf{h}$, which eventually approximates the actual log-marginal likelihood as follows, which gives the optimal inducing points to represent the approximated Gaussian distribution. 
\begin{flalign}
    L(\mathbf{x}_m) = log \Bigl(N\Bigl(\mathbf{y}|0, \sigma_n^2I+ k(\mathbf{x}, \mathbf{x})k(\mathbf{x}_m, \mathbf{x}_m)^{-1}k(\mathbf{x}_m,\mathbf{x})  -\frac{1}{2\sigma_n^2}Tr(k(\mathbf{x},\mathbf{x})-k(\mathbf{x}, \mathbf{x})k(\mathbf{x}_m\mathbf{x}_m)^{-1}k(\mathbf{x}_m, \mathbf{x})\Bigl)\Bigl) 
\end{flalign} Once we obtained the optimal values for $\mathbf{x}_m$, $\mu$ and $A$, which are the parameters of $\Phi(\mathbf{g}_m)$, can be determined  as 
\begin{equation*}
    \begin{aligned}
    \mu & = \frac{1}{\sigma_n^2} k(\mathbf{x}_m, \mathbf{x}_m)  \Bigl(k(\mathbf{x}_m, \mathbf{x}_m) + \sigma_n^{-2}k(\mathbf{x}_m, \mathbf{x})k(\mathbf{x}, \mathbf{x}_m)\Bigl)^{-1} k(\mathbf{x}_m,\mathbf{x}) \mathbf{y} ,\\
     A & = k(\mathbf{x}_m, \mathbf{x}_m) \Bigl(k(\mathbf{x}_m, \mathbf{x}_m) + \sigma_n^{-2}k(\mathbf{x}_m, \mathbf{x})k(\mathbf{x}, \mathbf{x}_m)\Bigl)^{-1} k(\mathbf{x}_m, \mathbf{x}_m),
\end{aligned} 
\end{equation*} respectively.
\begin{figure}[h!]
    \centering
    \includegraphics[width=0.4\linewidth]{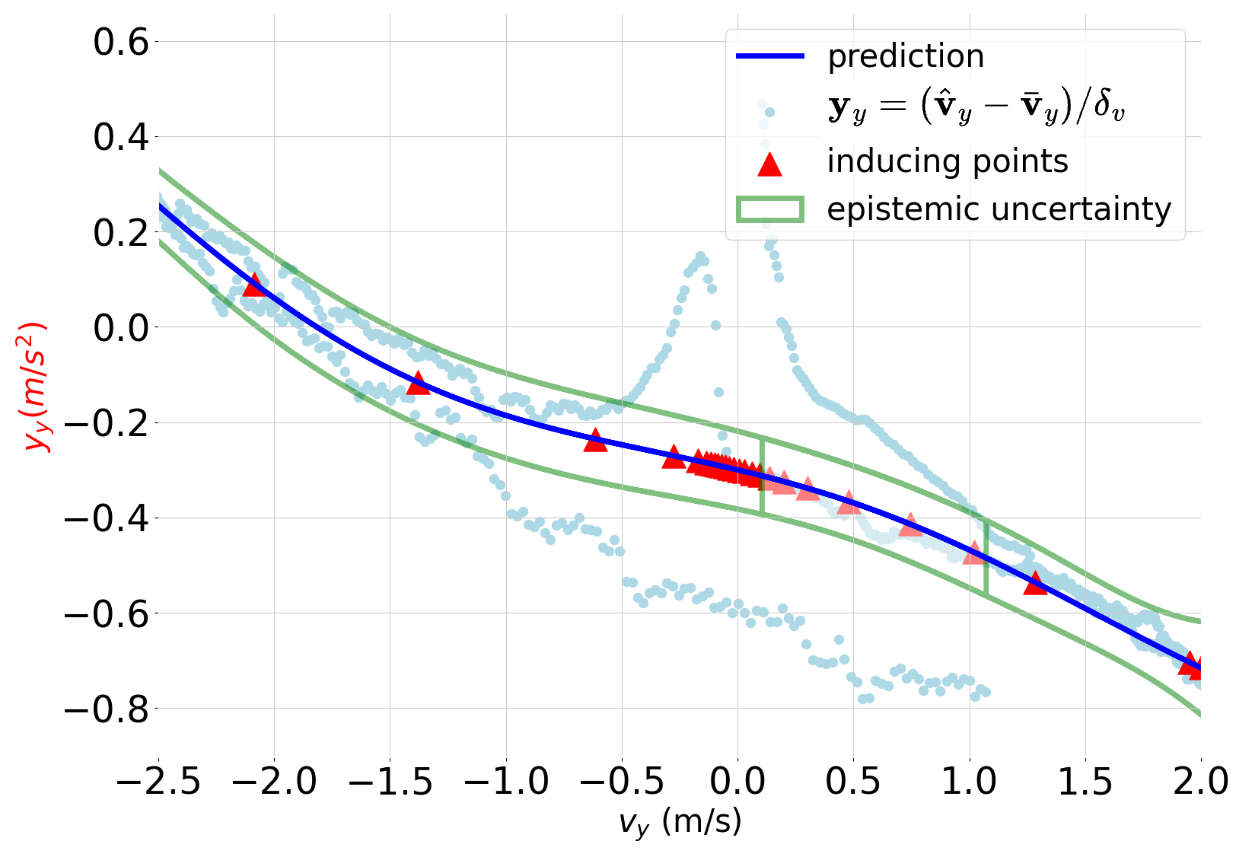}
    \caption{Epistemic uncertainty propagation of residual dynamics  $\mathbf{y}_y$ for changing of input velocity $\mathbf{v}_y$, where subscript $_y$ indicates the y axis. Variable $\mathbf{y}_y$, defined the rate of the different between predicted velocity $\bar{\mathbf{v}}_y$ and actual velocity $\hat{\mathbf{v}}_y$. Inducing points provide an approximation for $\mathbf{y}_y$. A Gaussian process is formed using such inducing points, i.e., sparse Gaussian process, which can be used to predict residual dynamics for corresponding input velocity $\mathbf{v}_y$.}
    \label{fig:induced_points}
\end{figure}After selecting the inducing points $\mathbf{x}_m$, only those points are considered to formulate the $\mathbf{g}(\mathbf{z})$ (eq.\ref{eq:estimated_model}). Afterwards, NMPC is formulated (eq.\ref{eq:nmpc}) with an augmented residual dynamics model.

\section*{Experimental procedure and results}

To assess the improvement of incorporating the proposed GP-based residual dynamics modelling into the local planner, we conducted several simulated and real-world experiments. For real-world experiments, the DJI M100 quadrotor was employed. The developed software stack is built on the ROS1 framework and implemented in C++. Since the local planner generates velocity and angular velocity commands, we employed the velocity control interface provided by the DJI Onboard-SDK-ROS\cite{dji}.  DJI control interface can handle control commands at a frequency of 20 Hz or higher. To bridge the gap between the DJI quadrotor and the local planner, we implemented a PD regulator~\cite{kulathunga2021trajectory} that ensures smooth command transmission at 30 Hz.


\subsection*{Collecting testing dataset}
\begin{figure}[h!]
    \centering
     \includegraphics[width=0.46\linewidth]{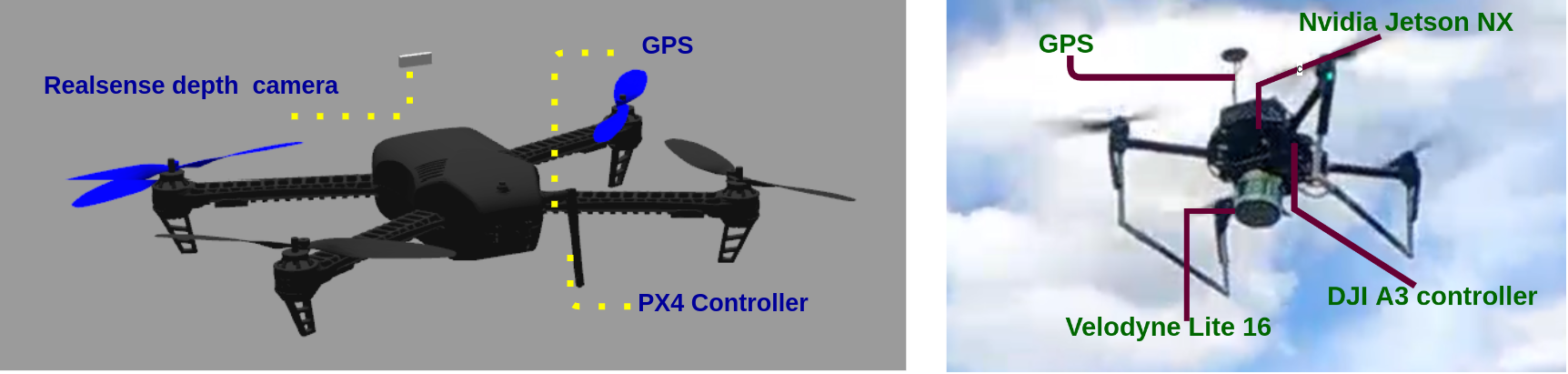}
    \includegraphics[width=0.46\linewidth]{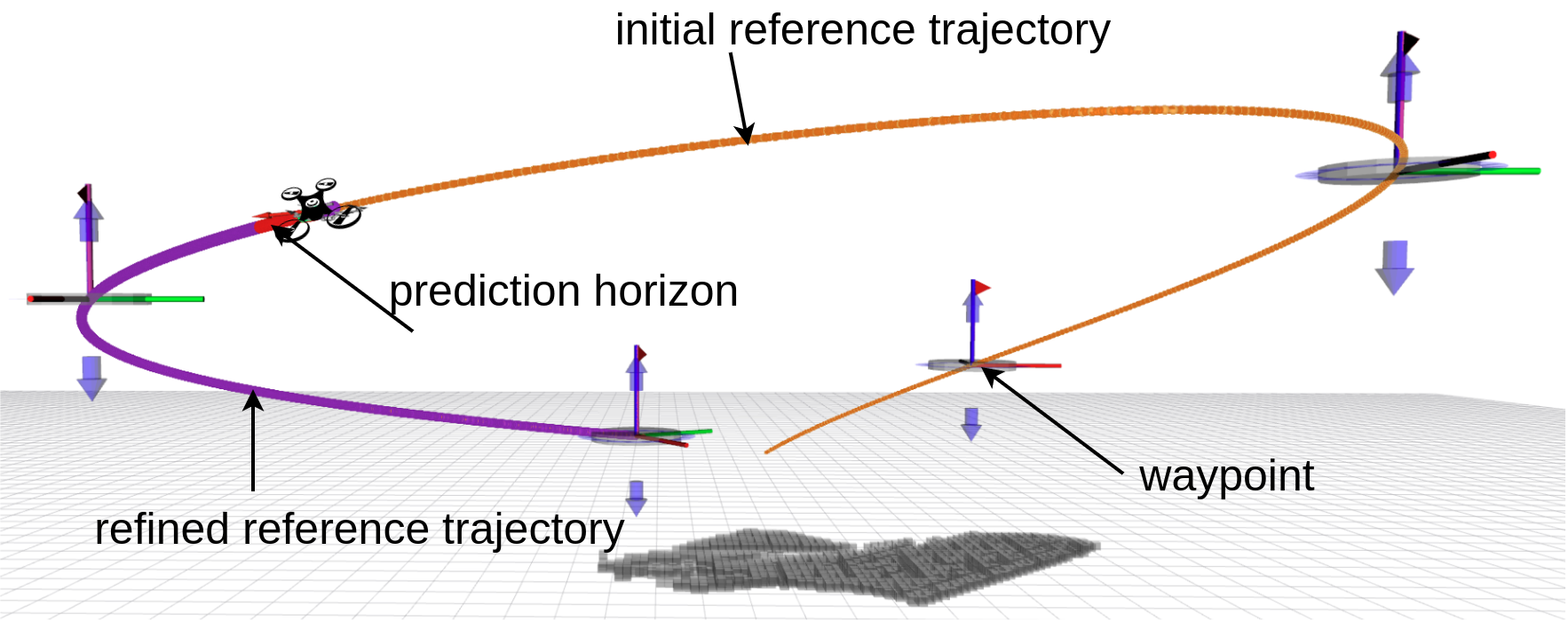}
    \caption{The right sub-figure: A trajectory of motion along all the axes was used to collect the data for learning and testing the latent representation of residual dynamics (Fig.~\ref{fig:induced_points}). The left sub-figure: DJI M100 quadrotor is used for real-world experiments and PX4-enabled quadrotor is used for simulated environments}
    \label{fig:sim_induced_points}
\end{figure}

Initially, twenty distinct trajectories were generated, as shown in Fig.~\ref{fig:sim_induced_points}. The objective was to collect a testing dataset in two forms: with and without obstacles. The testing dataset is structured as follows, which is similar to the training dataset: $D = \{X, \mathbf{y}\} = \{(\bar{\mathbf{v}}_i, \hat{\mathbf{v}}_i, \mathbf{\delta}_i), \mathbf{y}_i\}, i=0,...,n,$ where n represents the number of data points. In this structure:
$\bar{\mathbf{v}}_i, \hat{\mathbf{v}}_i, \mathbf{y}_i \in \mathbb{R}^3$ and $\delta_i \in \mathbb{R}$ represent the estimated velocity by NMPC, the actual velocity after applying the estimated velocity, the velocity residual, and the time difference between two consecutive data points, respectively. The terms $\bar{\mathbf{v}}_i, \hat{\mathbf{v}}_i$, and $\mathbf{y}_i$ are expressed as: $\bar{v}_{\mu}, \hat{v}_{\mu}, y_{\mu}, \; \mu \in \{x,y,z\},$ where $y_{\mu}$ and $v_\mu$ denote the expected residual dynamics and velocity component in the $\mu$ direction, respectively. These values are calculated at time $\mathbf{\delta}_i$ when using only the nominal model ($\mathbf{f}_{norm}$). For simplicity, $\mathbf{\delta}_i$ is defined as $\mathbf{\delta}_v$. Accordingly, the residual dynamics of the actual system are given by: $\mathbf{y}_i = (\hat{\mathbf{x}}_i - \bar{\mathbf{x}}_i)/\delta_i \rightarrow (\hat{\mathbf{v}}_i - \bar{\mathbf{v}}_i)/\delta_v$. The data were acquired using both a PX4-enabled quadrotor in a Gazebo-based simulated environment and a DJI M100 quadrotor (Fig.~\ref{fig:sim_induced_points}) in real-world conditions.
Since residual dynamics is invariant to the geometric representation of trajectory but variant to velocity changes, we focused on collecting the dataset such that the dataset represents the whole distribution of the velocity changes. Additionally, since we used two different quadrotors for simulated and real-world experiments, residual dynamics learning was carried out separately for both cases. However, for simplicity, we explained the results and finding considering the real-world setup. 

\subsection*{Modeling residual dynamics distribution}
In cluttered environments, the desired velocity is automatically reduced due to obstacle constraints and computational limitations, both of which are addressed by the local planner. Consequently, in such environments, accurate residual dynamics estimation is more crucial for lower velocities than for higher velocities. This is because the velocity profile typically falls within a lower range, such as 0 to 2 meters per second, particularly in cluttered environments with limited sensing capabilities. Hence, $g(\mathbf{z})$(eq.\ref{eq:gaussian_model_}) was modelled such that it could generate residual dynamics with low uncertainty for such velocities, i.e., by selecting more inducing points closer to lower velocity regions.
\begin{figure}[h!]
    \centering
    \includegraphics[width=0.40\linewidth]{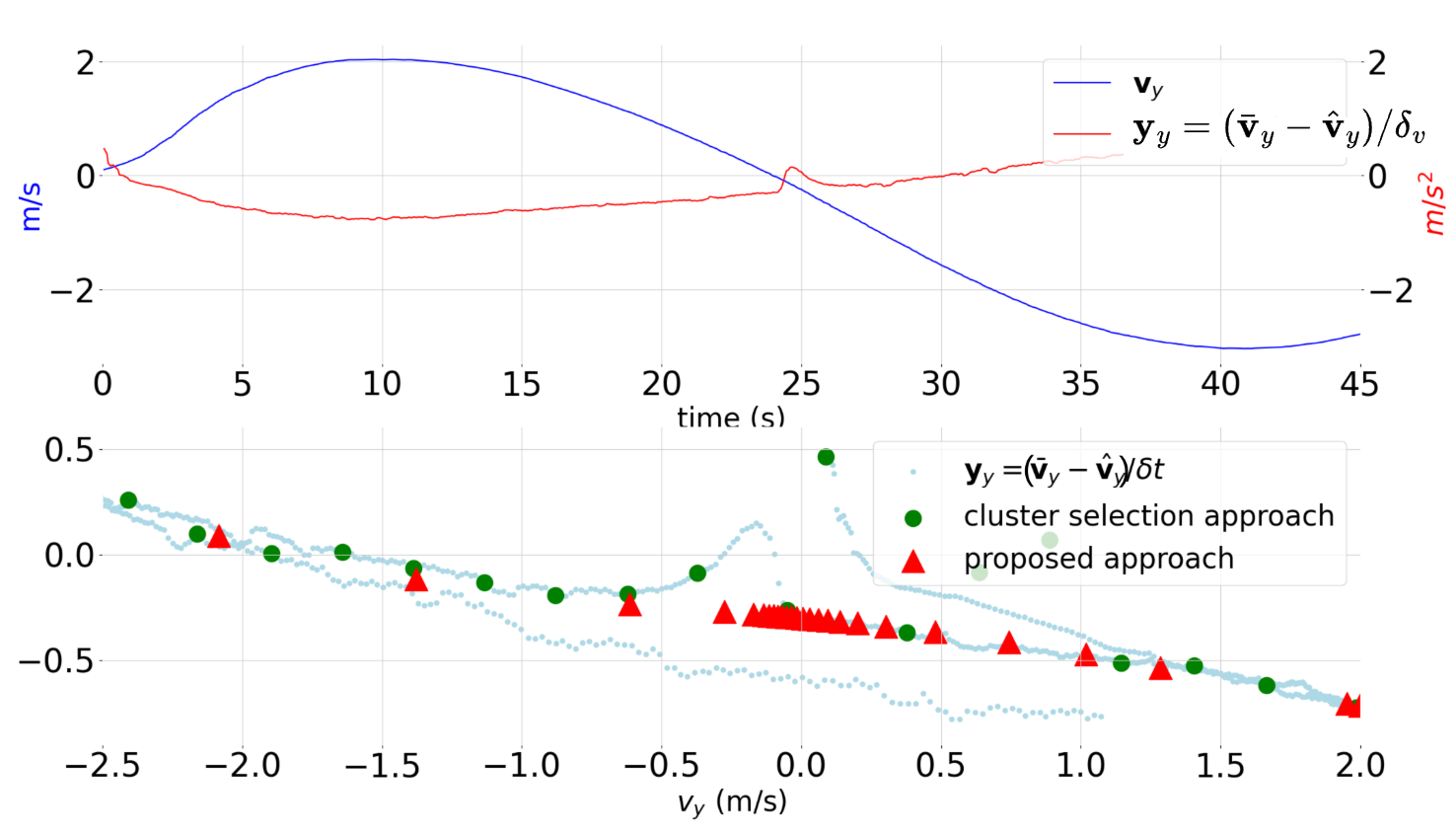}
    \includegraphics[width=0.56\linewidth]{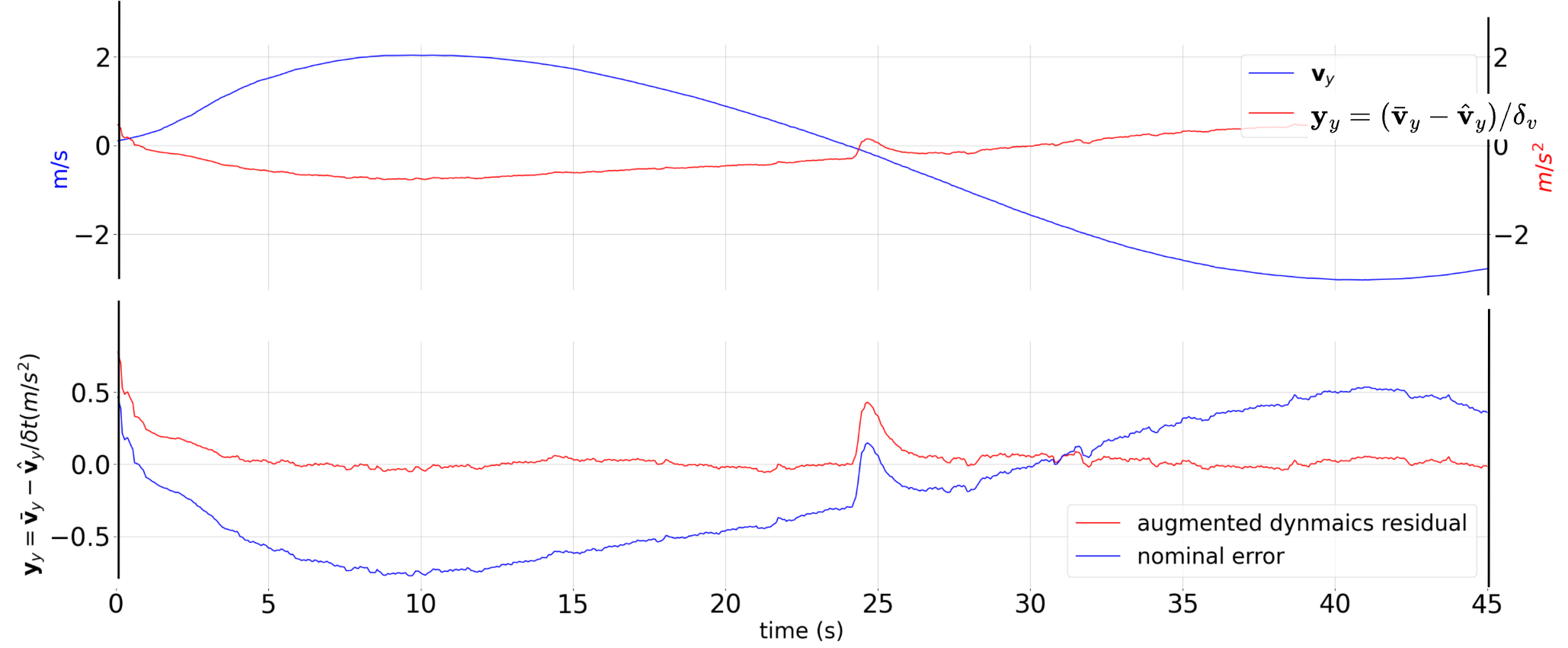}
    \caption{The relationship between residual dynamics $\mathbf{y}_y$ and input velocity $\mathbf{v}_y$. In left sub-figure: the first plot displays how $\mathbf{y}_y$ and $\mathbf{v}_y$ vary over time, while the second plot shows the approximation of the $\mathbf{y}_y$ using two different approaches: Sparse Gaussian Process-based (proposed) and cluster selection approach~\cite{torrente2021data}, in right sub-figure: residual dynamics before and after incorporating the learnt Sparse Gaussian process. Root Mean Square Error drops from 0.4012 to 0.0751 after introducing the residual dynamics into the nominal motion model}
    \label{fig:residual_dynamics_changes}
\end{figure} We have compared two inducing points selection approaches, i.e., the proposed approach and the cluster selection approach~\cite{torrente2021data}, for capturing residual dynamics along each axis: $\mathbf{y}_x, \mathbf{y}_y, $ and $\mathbf{y}_z$). The results of our experiments are given concerning the $y$ axis, i.e., $\mathbf{y}_y$. However, we applied the same procedure to other axes as well. The latter approach targets high-speed maneuvering provided that inducing points should capture the whole distribution, which is shown in Fig.~\ref{fig:residual_dynamics_changes} in green dots. In contrast, the proposed approach selects more inducing points (pink triangles) towards lower velocity ranges.  

\subsection*{Assessing the effect of residual dynamics produces on the nominal dynamics}
To assess which effect residual dynamics produces on the nominal dynamics eq.(\ref{eq:estimated_model}), we used the testing dataset that was acquired without considering obstacles  ($\{\mathbf{y}_{\mu},\mathbf{v}_{\mu}, \mathbf{\delta}_v\}$). Let $\bar{\mathbf{y}}_\mu = \mathbf{g}(\mathbf{v}_{\mu})$ be the expected residual dynamics. Then, the nominal error and the augmented residual error are fully determined by $\mathbf{y}_{\mu} \cdot \mathbf{\delta}_v$ and $(\mathbf{y}_{\mu} -\bar{\mathbf{y}}_\mu) \cdot \mathbf{\delta}_v$. We observed, that the accuracy of the augmented residual dynamics model has increased drastically compared to the standalone nominal model (Fig.~\ref{fig:residual_dynamics_changes}). Such an improvement helps to significantly reduce the trajectory tracking error since the control policy that is generated by NMPC after incorporating augmented residual dynamics can cope with the mean of epistemic uncertainty, residual dynamics, robustly. 

\subsection*{Selecting an optimal number of inducing points}
\begin{figure}[h!]
    \centering
    \includegraphics[width=0.6\linewidth]{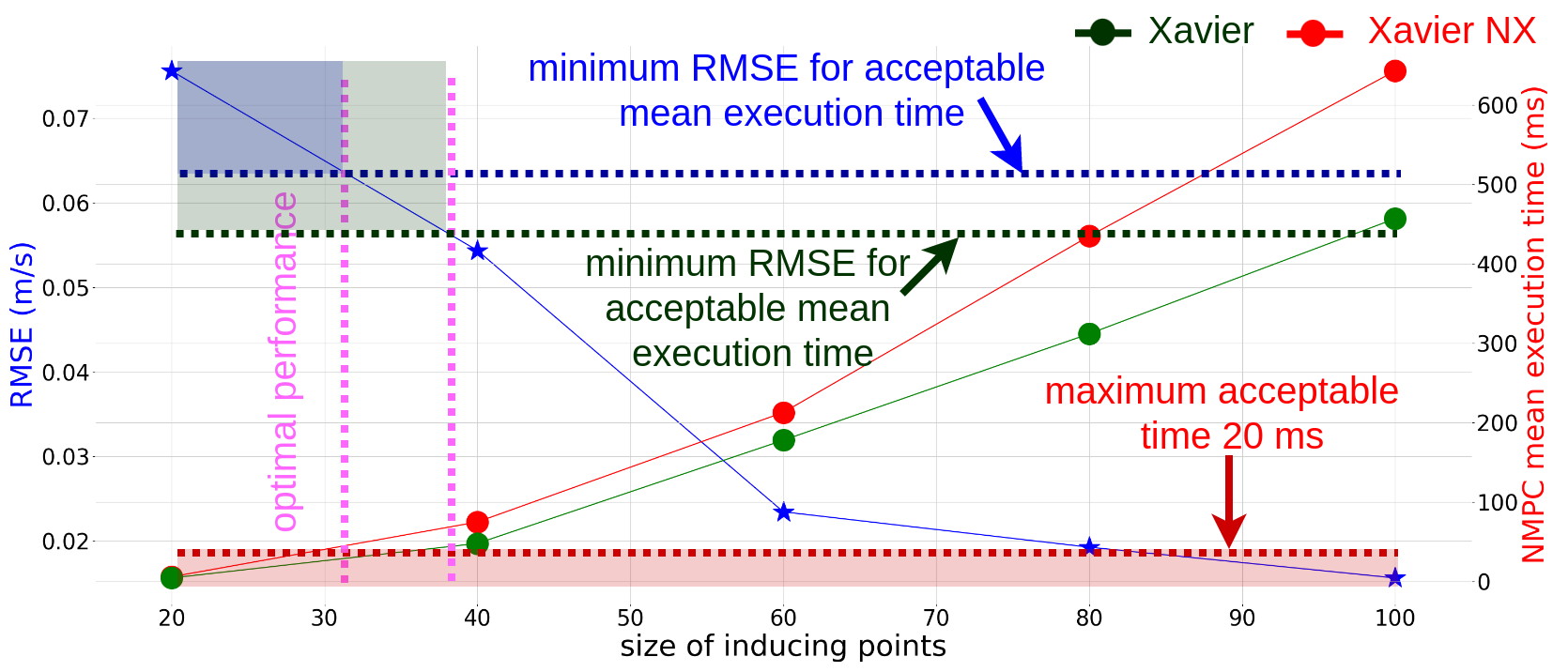}
    \caption{Correlation between the number of inducing points and computational power. The number of inducing points is inversely proportional to augmented dynamics residual RMSE, yet proportional to computational power for two embedded computers: Nvidia Xavier and Nvidia Xavier NX. Since RMSE shows no significant change compared to computation time, we selected 30 as the number of inducing points for formulating the Sparse Gaussian Process}
    \label{fig:points_and_execution_time}
\end{figure}

The execution time that is taken to estimate control policy is generated based on NMPC for maneuvering the quadrotor. This time is crucial for a smooth flight experience. We have experimented to find the correlation among the selected number of inducing points, NMPC execution time on two different embedded devices, and model accuracy (Fig.\ref{fig:points_and_execution_time}). As a result, we have selected only 30 inducing points to represent the residual dynamics. Such many inducing points were selected mainly due to two reasons.
 First, the accuracy improvement rate is almost negligible compared to NMPC execution time which grows exponentially when the number of inducing points is increased. Second, the Nvidia Xavier NX embedded computer was used for onboard computation. Moreover, there is no considerable advantage to using Nvidia Xavier over Nvidia Xavier NX since the optimization problem, i.e., local planner, is solved sequentially. The residual dynamics model learning is performed as an offline process, ensuring it does not impact real-time performance considerably. Therefore, the Graphics Processing Unit (GPU) is only utilized during the learning stage. Once the residual dynamics are estimated as a latent distribution, the model parameters are incorporated into the nominal model (eq.~\ref{eq:nmpc}). We employed Casadi~\cite{andersson2019casadi} with IPOPT~\cite{biegler2009large} as the primary solver for the local planner. Therefore, employing a large training dataset that accurately captures the residual dynamics distribution and carefully selecting optimal inducing points to represent the residual dynamics distribution effectively minimizes the residual dynamics of the model while having a minimum burden on computational cost.

\subsection*{Estimating root mean square error considering with and without obstacles}
Furthermore, we have calculated root mean square error (RMSE) with and without considering obstacles (Table.~\ref{tab:obs_no_obs}) in a simulated environment. The control policy generation can differ when considering obstacle constraints because NMPC can generate different control policies over time for different obstacle constraints. Thus, we kept the same initial reference trajectory running the planner with and without considering obstacles and then estimated the RMSE of residual dynamics.  
\begin{table}[h!]
  \centering
\caption{Comparison of how different types of inducing points selection affect the nominal model error} 
\label{tab:obs_no_obs}
\begin{tabular}{|llll|}
\hline
\begin{tabular}[c]{@{}l@{}}RMSE of augmented  residual  dynamics (m/s)\end{tabular} & \begin{tabular}[c]{@{}l@{}}The proposed approach\end{tabular} & \begin{tabular}[c]{@{}l@{}}Cluster selection approach\end{tabular} & \begin{tabular}[c]{@{}l@{}}Nominal error \end{tabular} \\ \hhline{|====|}
\begin{tabular}[c]{@{}l@{}}Without obstacles\end{tabular}  &  0.0849 &  0.1043  &  0.2116 \\ \hline
\begin{tabular}[c]{@{}l@{}}With obstacles\end{tabular} &  0.1972 & 0.2419  &  0.3314 \\ \hline
\end{tabular}
\end{table} RMSE is less when the augmented residual dynamics model is applied in obstacle-free environments, compared to obstacle-cluttered environments. Such behaviour is mainly due to the lack of stability of the global map, which is built incrementally as a result of updates within a small range, e.g., 5m from the quadrotor pose. Hence, the replanner iteratively refines the trajectory when the global map is changed. Thus, the local planner has to incorporate more obstacles and constraints from time to time. Such behaviour can lead to the trap of the quadrotor in local minima. None of such problems exists in obstacle-free environments.   

\subsection*{Avoiding trapping the quadrotor in local minima}\label{sec:local_minima}
\begin{figure}[h!]
    \centering
    \includegraphics[width=0.5\linewidth]{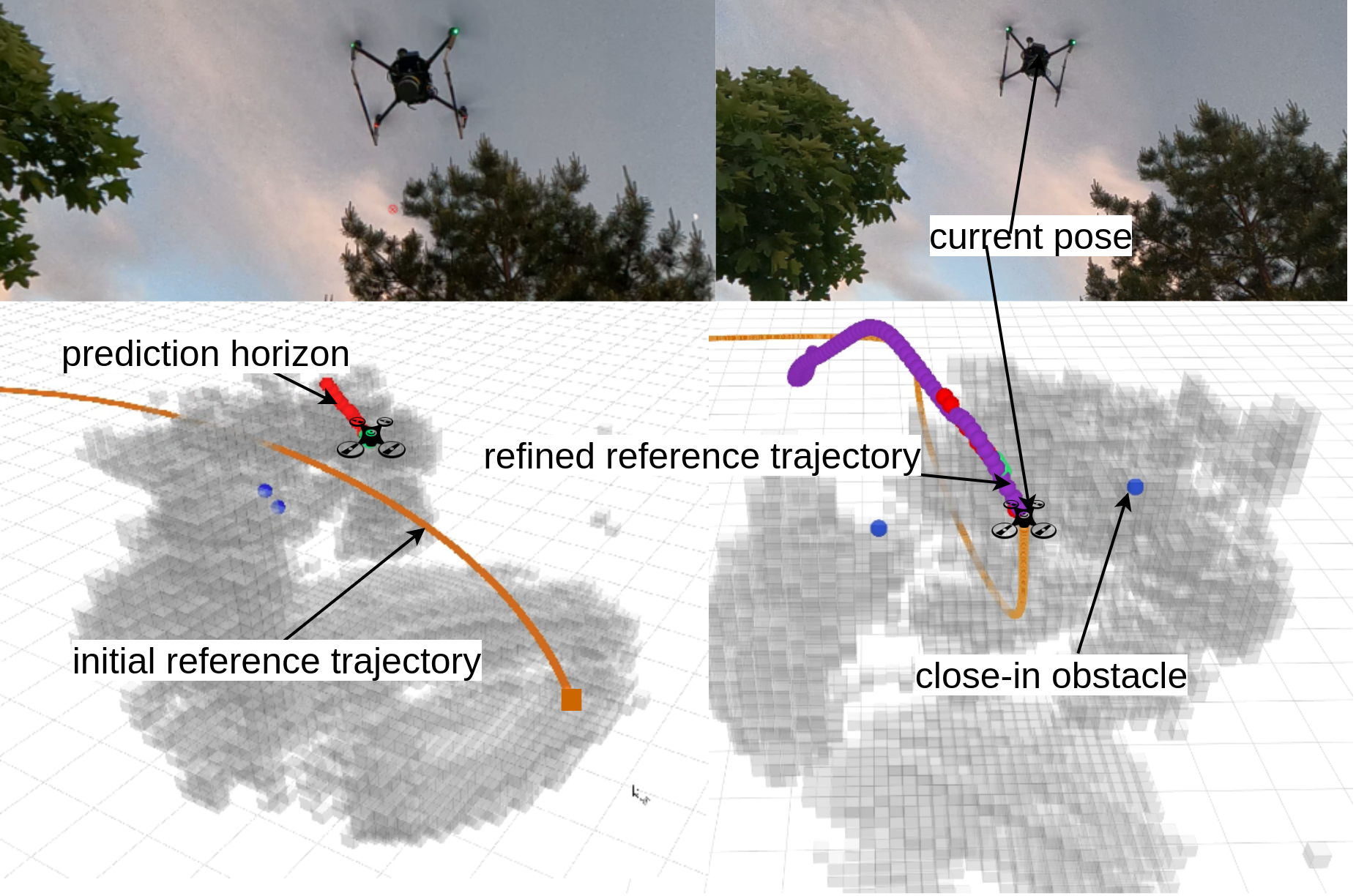}
    \caption{When the initial reference trajectory and the current pose of the quadrotor are far away from each other (left-side figure), e.g., 2m, regenerate the reference trajectory such that the replanner can try to push the reference trajectory further (the right-side figure)}
    \label{fig:local_minima}
\end{figure}

The initial reference trajectory can be within the obstacle zones since the quadrotor has no prior knowledge about the environment beyond the sensing range. Hence, the quadrotor can be far from the reference trajectory in the presence of a huge obstacle through the reference trajectory (Fig.~\ref{fig:local_minima}). In such a situation, the replanner~\cite{kulathunga2022optimization} fails to refine the trajectory due to time constraints.
 To overcome this issue, we proposed to regenerate a reference trajectory provided that the replanner can push the reference trajectory further away from the obstacles. In Fig.~\ref{fig:local_minima}, one such experiment shows how the proposed approach worked in real-world conditions. 

\subsection*{Experimenting in real-world conditions}

\begin{figure}[h!]
    \centering
    \includegraphics[width=0.7\linewidth]{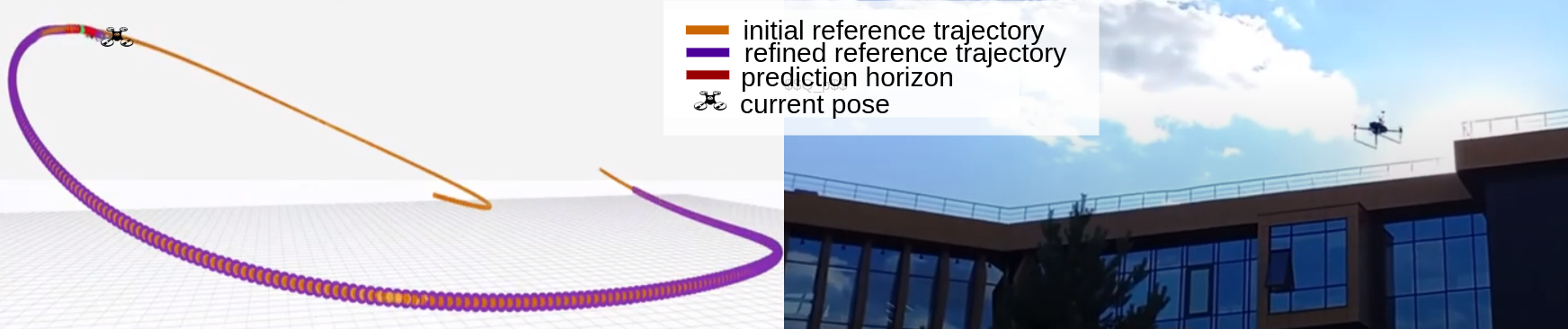}
    \caption{Example trajectory that is used to estimate the average RMSE of the augmented residual dynamics}
    \label{fig:experimentv00}
\end{figure}

To evaluate the proposed approach, we conducted two real-world experiments. In the first experiment, a set of twenty initial reference trajectories (Fig.~\ref{fig:experimentv00}) was generated for traversing an obstacle-free environment with a length range of 40 to 60 meters. The trajectories varied in their geometrical representations, while the maximum allowable velocity was set to 1.5 m/s. We first executed the experiment without considering residual dynamic estimation in the nominal model. Then, we repeated the experiment after incorporating residual dynamics. Both experiments were repeated three times for each trajectory to calculate the average RMSE of the augmented residual dynamics. Compared to the provided initial reference trajectory, the average RMSE for trajectories with and without augmented residual dynamics were 0.0851 and 0.2681, respectively. This indicates that incorporating residual dynamics reduced the residual dynamics error by almost half. The simulated experiment results (Table.~\ref{tab:obs_no_obs}) and real-world results show similar RMSE in obstacle-free environments. 

\begin{figure}[h!]
    \centering
    \includegraphics[width=1\linewidth]{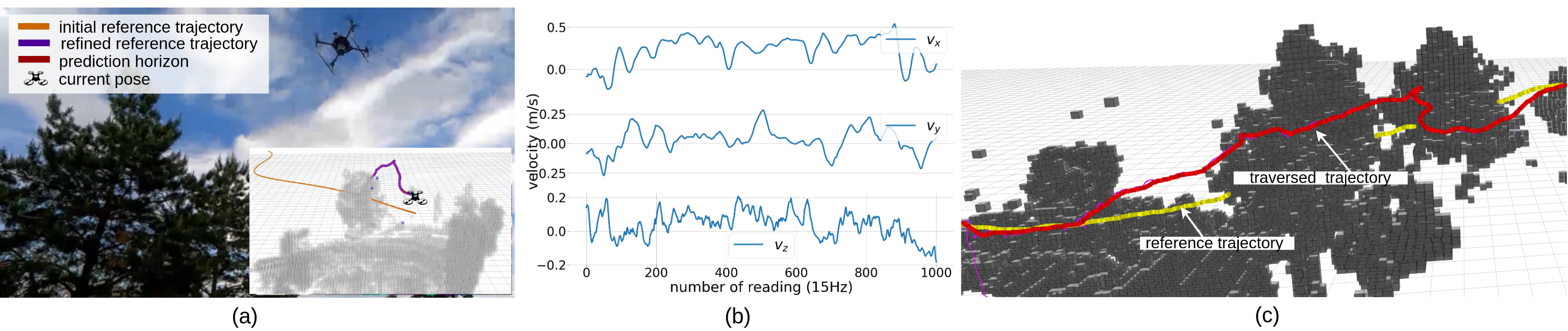}
    \caption{Experimental results for the proposed approach in a real-world setting. The initial reference trajectory within the obstacles is pushed iteratively. The distance between the refined trajectory and its closest obstacle poses is 1.5m. We set a 1.5 m distance between the refined trajectory (traversed trajectory) and the quadrotor pose. Due to computational constraints, the map of the environment is built 5m from the quadrotor pose incrementally, though LiDAR can measure up to 100m. Sub-figure (b) depicts the velocity profile for the experiment. Due to our reliance on GPS for quadrotor height estimation, the velocity along the z-axis exhibited significant instability compared to the other two axes.}
    \label{fig:experimentv1}
\end{figure}

In the second experiment, we only compared the RMSE of the trajectories after incorporating residual dynamics due to the dynamic nature of an environment with obstacles. The initial reference trajectory passes through a cluttered environment (Fig.~\ref{fig:experimentv1}, which is approximately 60m long. 
 The traversed trajectory distance is approximately 80m.The local planner we selected was specifically designed for low-speed maneuvering, as the primary objective was to navigate in cluttered environments. Consequently, we set the maximum velocity to 1.5 m/s. However, this restriction is not inherent to the proposed residual dynamics modeling approach. The initial reference trajectory was regenerated within the said traversed trajectory distance due to the local minima. Such an occasion can be seen in the provided video\cite{refvideo}. The RMSE of the augmented residual dynamics is 0.1018 (m/s). Even though it is hard to estimate the trajectory smoothness quantitatively, i.e., to assess control policy generation after introducing augmented residual dynamics, the smoothness has improved after incorporating augmented residual dynamics into the nominal model (Table. \ref{tab:obs_no_obs}).


\subsection*{Comparing with other similar recent works}
We compared the proposed approach with four different trajectory planners: sampling-based grid search followed by trajectory generation, NMPC-based replanner, hybrid planner (local and global), and KinoJGM. All the selected planners have their advantages and limitations with respect to low-speed maneuvers. The matrices that we considered are as follows: (1) mean computation time (MCT) (execution time for one iteration of the planner), (2) success rate (SR) (the number of times quadrotor reaches from start to goal without any collision), and (3) traversed distance (TD) (total distance of quadrotor that is traversed).
\begingroup
\setlength{\tabcolsep}{10pt} 
\renewcommand{\arraystretch}{1.7} 
\begin{table}[h!]
\centering
\caption{Comparative analysis in terms of goal-reaching. Ten different start and goal poses were considered whilst keeping the same environment}
\label{t:comparision}
\begin{tabular}{|llllll|}
\hline
\multirow{3}{*}{Algorithm} & \multirow{3}{*}{\begin{tabular}[c]{@{}l@{}}SF\end{tabular} } & \multirow{3}{*}{ \begin{tabular}[c]{@{}l@{}}MCT  (s)\end{tabular}} & \multicolumn{3}{l|}{Distance Estimation} \\ \cline{4-6} & &   & Mean (m)  & Max (m) & Min (m)        \\ \hhline{|=|=|=|=|=|=|}
\begin{tabular}[c]{@{}l@{}}\cite{kulathunga2020real}  RRT* max\_iterations=10000\end{tabular}            & 0.3      &   3.1 &    119.56   &  211.51  &   98.3\\ \hline         
\begin{tabular}[c]{@{}l@{}}\cite{kulathunga2021trajectory}  $P_{t_n}=15, u_r= 5m$\end{tabular}                         & 0.5 & 0.43 & 97.56  &   143.87  &  65.98 \\ \hline                      
\begin{tabular}[c]{@{}l@{}}\cite{kulathunga2021trajectory}  $P_{t_n}=30, u_r= 5m$ \end{tabular}                        & 0.6  & 0.49 &  95.34  &  139.6   & 76.51 \\ \hline
\begin{tabular}[c]{@{}l@{}}\cite{kulathunga2022optimization} $P_{t_n}=30, u_r= 5m$\end{tabular}                         & 0.9  & 0.5 &  93.14  &  145.22   & 79.4 \\ \hline
\begin{tabular}[c]{@{}l@{}}\cite{wang2022kinojgm}  KinoJSS $s_t=50 ms, P_{t_n}=20$ \end{tabular} & 0.7  & $\approx 0.008$ &  110.48  &  131.45   & 95.32 \\ \hline
\begin{tabular}[c]{@{}l@{}}\cite{wang2022kinojgm}  GP-MPC $s_t=50 ms, P_{t_n}=20$\end{tabular}   & 0.6  & $ \approx 0.021$  &  99.23  &  119.54   & 90.18 \\ \hline
\begin{tabular}[c]{@{}l@{}}\cite{wang2022kinojgm}  KinoJGM $s_t=50 ms, P_{t_n}=20$\end{tabular}  & 0.8    & $\approx 0.034$  &  90.34  &  120.53   & 81.37 \\ \hline
\begin{tabular}[c]{@{}l@{}}Proposed $P_{t_n}=10, u_r= 4m$, $x_m=20$ \end{tabular} & 0.9    & $0.03 \pm 0.01$  &  84.89 &  124.90   & 81.78 \\ \hline
\begin{tabular}[c]{@{}l@{}}Proposed $P_{t_n}=15, u_r= 6m$, $x_m=20$\end{tabular}  & \textbf{1.0}   & $\mathbf{0.04}\pm \textbf{0.01}$ &   85.41  &  115.89   & 79.54 \\ \rowcolor{LightCyan} \hline
\end{tabular}
\begin{tablenotes}
     \item[1]  SF: Success Fraction, MCT: Mean Computation Time, $P_{t_n}$: MPC prediction horizon length, $u_r$: map update range incrementally, $x_m$: number of inducing points, $s_t$: sampling time, KinoJSS; GP-MPC; KinoJGM are three different variants of~\cite{wang2022kinojgm} 
\end{tablenotes}
\end{table}
\endgroup

 In this experiment, we use a random forest  with a density of (60m$\times$60m$\times$10m). After that, we selected 10 different start and target poses and executed each planner. The experiment results are given in Table.~\ref{t:comparision}. For the sampling-based planner, MCT is high because it generates nondeterministic trajectories due to the nature of the planner. Replanner generates locally near-optimal control policies, yet execution time is high due to design constraints, i.e., the local planner is formulated as a constraint nonlinear optimization problem. Moreover, the quadrotor can trap in local minima since the planner optimizes locally. Compared to the listed two planners, the hybrid approach has high MCT and SR.


The latter planner executes two sub-planners, local and global, simultaneously. Hence, the planner observes the environment in the distance and generates a near-optimal control policy to reach the goal. However, since a simplified motion model was used to generate near-optimal control policy, dynamics uncertainty (or residual dynamics) between desired control that the planner generated and the actual control that the quadrotor obtained is considerably high. Such behaviour leads to trapping the quadrotor in local minima. The proposed approach is a modification of this planner by addressing two issues. Traps in local minima and residual dynamics are the two issues that the proposed planner addressed. Resolving those issues helps to obtain a low TD compared to the hybrid planner and a slightly higher SR. In addition, the proposed planner was compared with the KinoJGM planner and its variants: GP-MPC and KinoJSS. Those results were provided by the authors of KinoJGM.  GP-MPC planner is conceptually similar to the proposed one. However, it has no recovery mechanism and ways to avoid local minima unlike ours. These are the reasons the proposed planner achieved higher SR compared to GP-MPC. MCT of the proposed is slightly lower compared to GP-MPC. However, the obtained MCT is acceptable for low-speed maneuvers.

\section*{Discussion}

The trajectory planning problems in the plan-based control paradigm, in general, are solved adhering to these steps: path search~\cite{wang2023learning}, initial trajectory generation and trajectory refinement~\cite{liu2022rapid}, high-level control command generation that can be achieved by several approaches, including differential flatness mapping~\cite{talke2022autonomous}, receding horizon planning, and finally low-level control commands generation using a flight controller, for example, PX4, DJI. These flight controllers operate independently irrespective of high-level planners. Moreover, due to their independence, such controllers reduce the overhead and complexity of developing high-level planning algorithms. In other words, the same planner can be deployed on different firmware by implementing an interface between a high-level planner and a low-level controller. Therefore, residual dynamics arise between the high-level planner and the low-level controller. Gaussian Process-based techniques are used to model these residual dynamics in the recent past not only for MAVs but also other types of vehicles~\cite{narayanan2023physics}, \cite{xu2023data}.

In recent years, MAVs-related manifestations,  e.g., trajectory tracking, exploration, expanding the range for venturing out computationally expensive techniques in various disciplines, including agriculture, aerial photography, and crop monitoring~\cite{fikri2023review},\cite{nduku2023global}. Low-speed maneuvering, in general, is preferred over high-speed maneuvering in executing such demanding tasks due to task complexity. An example of such low-speed maneuver need is trajectory planning, where the environment is obstacle-rich. Therefore, the kinematic modelling of a quadrotor was considered since the scope of this work is for low-speed maneuvers without consideration of dynamic effects (external or/and internal). Similar assumptions were used in~\cite{anderegg2023farm} for weed detection in crops, which showed the necessity of low-speed maneuvering. On the other hand, several studies have been carried out for high-speed maneuvers~\cite{rojas2021board}, \cite{song2020learning}, \cite{8424034} with consideration of residual dynamics estimation, while most of them struggle with high-computational demands for on-board processing and difficulty in maneuvering in cluttered environments. The proposed framework can be adapted for high-speed maneuvers as well. However, it requires adding a behavioural planner that switches between planners according to the requirements, for example, in the cluttering environment we can switch to the low maneuvers while keeping the same control command generation. That will give us several benefits compared to existing solutions, e.g.,  residual dynamics models can be trained separately for low-speed maneuvers and high-speed maneuvers. Hence, the proposed solution can be adapted to general control with high and low-speed maneuvers introducing two or more operation modes and switching between them based on the external conditions.

Since Sparse Gaussian Processing is a non-parametric model, it does not require any parameter tuning.  However, it is necessary to provide a proper training set that can capture the whole distribution of the latent space, i.e., residual dynamics. The simulated experiments of the proposed approach were carried out with an IRIS quadrotor with a PX4 flight controller. The real-world experiments were carried out with a DJI M100 with an A3 flight controller. In both cases, the initial step was to fine-tune the hyper-parameters of the PD regulator. This process is required before running the proposed motion planner. Suppose the proposed approach is deployed in another MAV. In that case, it is required to fine-turn the listed parameters since the high-level planner does not know any information about the low-level controller. However, this is a one-time process. Recently, auto PID tuning approaches were proposed~\cite{stamate2023improvement}. Hence, it is required to investigate these aspects to improve the parameter tuning of the proposed approach.   


The source code and complete experiments are available at Github\cite{refvideo}

\bibliography{sample}

\section*{Acknowledgement}

This work was supported by the InnovateUK-funded project Agri-OpenCore [grant number 10041179]


\section*{Data Availability}
The datasets generated and/or analysed during the current study are available in the trajectory-tracker repository, \url{https://github.com/GPrathap/trajectory-tracker}

\section*{Additional information}

\textbf{Competing interests}. 

\textbf{Financial competing interests}

No, I (Geesara Kulathunga (gkulathunga@lincoln.ac.uk)) declare the authors have no competing interests as defined by Nature Research, or other interests
that might be perceived to influence the interpretation of the article.

\textbf{Non-financial competing interests}

No, (Geesara Kulathunga (gkulathunga@lincoln.ac.uk)) declare the authors have no non-financial competing interests as defined by Nature Research,
or other interests that might be perceived to influence the interpretation of the article.


\end{document}